\documentclass[cmc,article,accept,moreauthors,pdftex]{Definitions/tsp} 
\usepackage[normalem]{ulem} 
\usepackage{amsfonts} 
\usepackage{mathcomp} 
\usepackage{CJKutf8} 
\usepackage{pifont} 
\usepackage{bm} 
\usepackage{bbm} 
\graphicspath{{./Definitions/}} 

\makeatletter
\def\T@n@@nc@d@ngM@cr@M@d{}
\def\LY@n@@nc@d@ngM@cr@M@d{}
\makeatother

\let\orignewcommand\newcommand  
\let\newcommand\providecommand  
\usepackage{verse}
\let\newcommand\orignewcommand  

\newsavebox\foobox

\renewcommand{\dagger}{\mathchar"2279}
\renewcommand{\ddagger}{\mathchar"227A}

\newcommand{\mmathit}[1]{
  \ifthenelse{\equal{#1}{\ln}}{\mathit{ln}}{
    \ifthenelse{\equal{#1}{\max}}{\mathit{max}}{\mathit{#1}}
  }
}
\makeatother
\robustify{\footnote}

\DeclareUnicodeCharacter{1E45}{\.{n}}
\DeclareUnicodeCharacter{1E41}{\.{m}}
\DeclareUnicodeCharacter{2003}{\quad}
\DeclareUnicodeCharacter{2009}{\thinspace}
\DeclareUnicodeCharacter{2002}{\enspace{}}
\DeclareUnicodeCharacter{2005}{\thinspace}
\DeclareUnicodeCharacter{0263}{\textipa{G}}
\DeclareUnicodeCharacter{A0}{~}
\DeclareUnicodeCharacter{2460}{\textcircled{\scriptsize{1}}}
\DeclareUnicodeCharacter{2461}{\textcircled{\scriptsize{2}}}
\DeclareUnicodeCharacter{2462}{\textcircled{\scriptsize{3}}}
\DeclareUnicodeCharacter{2463}{\textcircled{\scriptsize{4}}}
\DeclareUnicodeCharacter{2464}{\textcircled{\scriptsize{5}}}
\DeclareUnicodeCharacter{2465}{\textcircled{\scriptsize{6}}}
\DeclareUnicodeCharacter{2466}{\textcircled{\scriptsize{7}}}
\DeclareUnicodeCharacter{2467}{\textcircled{\scriptsize{8}}}
\DeclareUnicodeCharacter{2468}{\textcircled{\scriptsize{9}}}
\DeclareUnicodeCharacter{2070}{\textsuperscript{0}}
\DeclareUnicodeCharacter{2074}{\textsuperscript{4}}
\DeclareUnicodeCharacter{2075}{\textsuperscript{5}}
\DeclareUnicodeCharacter{2076}{\textsuperscript{6}}
\DeclareUnicodeCharacter{2077}{\textsuperscript{7}}
\DeclareUnicodeCharacter{2078}{\textsuperscript{8}}
\DeclareUnicodeCharacter{2079}{\textsuperscript{9}}
\DeclareUnicodeCharacter{02C2}{<}
\DeclareUnicodeCharacter{2033}{\relax\ifmmode '' \else $''$\fi}
\DeclareUnicodeCharacter{2034}{\relax\ifmmode ''' \else $'''$\fi}
\DeclareUnicodeCharacter{2026}{\relax\ifmmode … \else $\ldots$\fi}
\DeclareUnicodeCharacter{0229}{\c{e}}
\DeclareUnicodeCharacter{016F}{\r{u}}
\DeclareUnicodeCharacter{127}{\relax\ifmmode\rm\hbar\else $\rm\hbar$\fi}
\DeclareUnicodeCharacter{3AC}{\relax\ifmmode\acute{\alpha}\else $\acute{\alpha}$\fi}
\DeclareUnicodeCharacter{3AD}{\relax\ifmmode\acute{\varepsilon}\else $\acute{\varepsilon}$\fi}
\DeclareUnicodeCharacter{3AE}{\relax\ifmmode\acute{\eta}\else $\acute{\eta}$\fi}
\DeclareUnicodeCharacter{3AF}{\relax\ifmmode\acute{\iota}\else $\acute{\iota}$\fi}
\DeclareUnicodeCharacter{3CC}{\relax\ifmmode\acute{o}\else $\acute{o}$\fi}
\DeclareUnicodeCharacter{3CD}{\relax\ifmmode\acute{\upsilon}\else $\acute{\upsilon}$\fi}
\DeclareUnicodeCharacter{3CE}{\relax\ifmmode\acute{\omega}\else $\acute{\omega}$\fi}
\DeclareUnicodeCharacter{391}{A}
\DeclareUnicodeCharacter{392}{B}
\DeclareUnicodeCharacter{395}{E}
\DeclareUnicodeCharacter{396}{Z}
\DeclareUnicodeCharacter{397}{H}
\DeclareUnicodeCharacter{399}{I}
\DeclareUnicodeCharacter{39A}{K}
\DeclareUnicodeCharacter{39C}{M}
\DeclareUnicodeCharacter{39D}{N}
\DeclareUnicodeCharacter{39F}{O}
\DeclareUnicodeCharacter{3A1}{P}
\DeclareUnicodeCharacter{3A4}{T}
\DeclareUnicodeCharacter{3A7}{X}

\DeclareUnicodeCharacter{27E6}{\relax\ifmmode \llbracket \else $\llbracket$\fi}
\DeclareUnicodeCharacter{27E7}{\relax\ifmmode \rrbracket \else $\rrbracket$\fi}
\DeclareUnicodeCharacter{00B5}{\relax\ifmmode \mu{} \else $\mu$\fi}
\DeclareUnicodeCharacter{1D434}{\relax\ifmmode A \else $A$\fi}
\DeclareUnicodeCharacter{1D435}{\relax\ifmmode B \else $B$\fi}
\DeclareUnicodeCharacter{1D436}{\relax\ifmmode C \else $C$\fi}
\DeclareUnicodeCharacter{1D437}{\relax\ifmmode D \else $D$\fi}
\DeclareUnicodeCharacter{1D438}{\relax\ifmmode E \else $E$\fi}
\DeclareUnicodeCharacter{1D439}{\relax\ifmmode F \else $F$\fi}
\DeclareUnicodeCharacter{1D43A}{\relax\ifmmode G \else $G$\fi}
\DeclareUnicodeCharacter{1D43B}{\relax\ifmmode H \else $H$\fi}
\DeclareUnicodeCharacter{1D43C}{\relax\ifmmode I \else $I$\fi}
\DeclareUnicodeCharacter{1D43D}{\relax\ifmmode J \else $J$\fi}
\DeclareUnicodeCharacter{1D43E}{\relax\ifmmode K \else $K$\fi}
\DeclareUnicodeCharacter{1D43F}{\relax\ifmmode L \else $L$\fi}
\DeclareUnicodeCharacter{1D440}{\relax\ifmmode M \else $M$\fi}
\DeclareUnicodeCharacter{1D441}{\relax\ifmmode N \else $N$\fi}
\DeclareUnicodeCharacter{1D442}{\relax\ifmmode O \else $O$\fi}
\DeclareUnicodeCharacter{1D443}{\relax\ifmmode P \else $P$\fi}
\DeclareUnicodeCharacter{1D444}{\relax\ifmmode Q \else $Q$\fi}
\DeclareUnicodeCharacter{1D445}{\relax\ifmmode R \else $R$\fi}
\DeclareUnicodeCharacter{1D446}{\relax\ifmmode S \else $S$\fi}
\DeclareUnicodeCharacter{1D447}{\relax\ifmmode T \else $T$\fi}
\DeclareUnicodeCharacter{1D448}{\relax\ifmmode U \else $U$\fi}
\DeclareUnicodeCharacter{1D449}{\relax\ifmmode V \else $V$\fi}
\DeclareUnicodeCharacter{1D44A}{\relax\ifmmode W \else $W$\fi}
\DeclareUnicodeCharacter{1D44B}{\relax\ifmmode X \else $X$\fi}
\DeclareUnicodeCharacter{1D44C}{\relax\ifmmode Y \else $Y$\fi}
\DeclareUnicodeCharacter{1D44D}{\relax\ifmmode Z \else $Z$\fi}
\DeclareUnicodeCharacter{1D44E}{\relax\ifmmode a \else $a$\fi}
\DeclareUnicodeCharacter{1D44F}{\relax\ifmmode b \else $b$\fi}
\DeclareUnicodeCharacter{1D450}{\relax\ifmmode c \else $c$\fi}
\DeclareUnicodeCharacter{1D451}{\relax\ifmmode d \else $d$\fi}
\DeclareUnicodeCharacter{1D452}{\relax\ifmmode e \else $e$\fi}
\DeclareUnicodeCharacter{1D453}{\relax\ifmmode f \else $f$\fi}
\DeclareUnicodeCharacter{1D454}{\relax\ifmmode g \else $g$\fi}
\DeclareUnicodeCharacter{1D456}{\relax\ifmmode i \else $i$\fi}
\DeclareUnicodeCharacter{1D457}{\relax\ifmmode j \else $j$\fi}
\DeclareUnicodeCharacter{1D458}{\relax\ifmmode k \else $k$\fi}
\DeclareUnicodeCharacter{1D459}{\relax\ifmmode l \else $l$\fi}
\DeclareUnicodeCharacter{1D45A}{\relax\ifmmode m \else $m$\fi}
\DeclareUnicodeCharacter{1D45B}{\relax\ifmmode n \else $n$\fi}
\DeclareUnicodeCharacter{1D45C}{\relax\ifmmode o \else $o$\fi}
\DeclareUnicodeCharacter{1D45D}{\relax\ifmmode p \else $p$\fi}
\DeclareUnicodeCharacter{1D45E}{\relax\ifmmode q \else $q$\fi}
\DeclareUnicodeCharacter{1D45F}{\relax\ifmmode r \else $r$\fi}
\DeclareUnicodeCharacter{1D460}{\relax\ifmmode s \else $s$\fi}
\DeclareUnicodeCharacter{1D461}{\relax\ifmmode t \else $t$\fi}
\DeclareUnicodeCharacter{1D462}{\relax\ifmmode u \else $u$\fi}
\DeclareUnicodeCharacter{1D463}{\relax\ifmmode v \else $v$\fi}
\DeclareUnicodeCharacter{1D464}{\relax\ifmmode w \else $w$\fi}
\DeclareUnicodeCharacter{1D465}{\relax\ifmmode x \else $x$\fi}
\DeclareUnicodeCharacter{1D466}{\relax\ifmmode y \else $y$\fi}
\DeclareUnicodeCharacter{1D467}{\relax\ifmmode z \else $z$\fi}

\DeclareUnicodeCharacter{1E67}{\.{\v s}}
\DeclareUnicodeCharacter{1E11}{\relax\ifmmode \c{d} \else $\c{d}$\fi}
\DeclareUnicodeCharacter{1ECB}{\relax\ifmmode \d{i} \else $\d{i}$\fi}
\DeclareUnicodeCharacter{1D8D}{\relax\ifmmode \textlhookx \else $\textlhookx$\fi}
\DeclareUnicodeCharacter{104}{\relax\ifmmode \k{A} \else $\k{A}$\fi}
\DeclareUnicodeCharacter{211E}{\relax\ifmmode \textrecipe \else $\textrecipe$\fi}
\DeclareUnicodeCharacter{29D}{\relax\ifmmode \textctj \else $\textctj$\fi}

\DeclareUnicodeCharacter{1E2E}{\'{\"I}}
\DeclareUnicodeCharacter{23F}{\textrts}

\DeclareUnicodeCharacter{2C73}{\varw}

\DeclareUnicodeCharacter{2127}{\mho}

\DeclareUnicodeCharacter{28C}{\textturnv}
\DeclareUnicodeCharacter{252}{\textturnscripta}
\DeclareUnicodeCharacter{259}{\schwa}
\DeclareUnicodeCharacter{25B}{\m{e}}
\DeclareUnicodeCharacter{266}{\m{h}}
\DeclareUnicodeCharacter{127}{\B{h}}
\DeclareUnicodeCharacter{27E}{\textfishhookr}
\DeclareUnicodeCharacter{281}{\textinvscr}

\continuouspages{}
\firstpage{1} 
\pubvolume{0}
\issuenum{0}
\elocationid{0}
\articlenumber{0}
\pubyear{2026}
\copyrightyear{2026}
\datereceived{Day Month Year}
\dateaccepted{Day Month Year}
\dateonlinefirst{}
\datepublished{Day Month Year}

\usepackage[OT1,OT2,T2A,T2B,T2C,T3,T5,T1]{fontenc}
\usepackage[russian,english]{babel}
\usepackage{xcolor}
\usepackage{algorithm}
\usepackage{algorithmicx}
\usepackage{algpseudocode}

\Title{pFedUL: Layer-Aware Federated Unlearning for Personalized Federated Learning}

\Author{Zhuodong Liu\textsuperscript{1}, Xiangyu Li\textsuperscript{2,*} and Zhihao Zhang\textsuperscript{1}}

\AuthorNames{Zhuodong Liu, Xiangyu Li and Zhihao Zhang}

\address{%
\textsuperscript{1}Saunders College of Business, Rochester Institute of Technology, Rochester, NY 14623, USA\\
\textsuperscript{2}School of Electrical and Computer Engineering, Georgia Institute of Technology, Atlanta, GA 30332, USA
}%

\corres{Corresponding Author: Xiangyu Li. Email: xli985@gatech.edu}

\abstract{Federated unlearning (FU) enables the removal of specific data contributions from federated learning (FL) models to comply with regulations such as the General Data Protection Regulation (GDPR). However, most existing FU methods are designed for the FedAvg paradigm, where all clients share a single global model. In practice, personalized federated learning (pFL) methods such as FedPer, FedRep, Ditto, and FedBN have become widely adopted due to their superior handling of non-IID data. These methods decompose the model into shared global layers and client-specific personalized layers, fundamentally altering the semantics of unlearning, yet this setting has received little attention. We formalize FU under the pFL paradigm, identifying a tension between unlearning completeness on shared layers and personalization preservation for remaining clients. We then propose pFedUL, a layer-aware selective unlearning framework comprising three components: (1) gradient-based layer-wise contribution attribution that separately quantifies the target client's influence on shared and personalized parameters, (2) adaptive selective unlearning that applies differentiated forgetting strategies across layer types, and (3) a lightweight recalibration protocol enabling remaining clients to restore personalization with minimal overhead. We further introduce two new metrics, Personalization Preservation Score (PPS) and Cross-client Fairness Index (CFI), to evaluate pFL-specific unlearning quality. Experiments on CIFAR-10, CIFAR-100, and FEMNIST under varying non-IID settings indicate that pFedUL achieves unlearning effectiveness comparable to full retraining while maintaining an average of 97.3\% personalized accuracy for remaining clients. Compared with six state-of-the-art FU methods adapted to the pFL setting, pFedUL consistently achieves superior personalization preservation, improving over the best existing method by 6.3\% in PPS on average with an 8.4$\times$ speedup, averaged across all tested pFL architectures and datasets.}


\begin{document}

\section{Introduction} \label{sect:s1}

Federated learning (FL) has emerged as a prominent distributed machine learning paradigm that enables multiple clients to collaboratively train a shared model without directly exchanging raw data \cite{ref-1}. By keeping training data locally on each client and communicating only model updates, FL provides inherent privacy advantages over centralized learning \cite{ref-2}. This paradigm has been widely deployed in domains such as healthcare, mobile keyboard prediction, and recommendation systems, where data sensitivity and regulatory compliance are of paramount importance \cite{ref-3}.

Despite its privacy-preserving design, FL does not fully satisfy the ``right to be forgotten'' mandated by data protection regulations such as the European General Data Protection Regulation (GDPR) \cite{ref-4} and similar data protection regulations worldwide. These regulations require that, upon a data owner's request, all traces of their data contributions must be effectively removed from the trained model. In centralized settings, machine unlearning has been extensively studied to address this challenge. Representative methods include SISA \cite{ref-6}, which partitions training data into disjoint shards and retrains only the affected subset, and influence-function-based approaches \cite{ref-7} that approximate the effect of removing specific data points. However, retraining from scratch in federated settings is prohibitively expensive due to communication overhead and the potential unavailability of participating clients \cite{ref-8}.

To address this challenge, federated unlearning (FU) has attracted increasing research attention \cite{ref-9}. FedEraser \cite{ref-10} pioneered this direction by leveraging stored historical gradient updates to calibrate the global model without full retraining. Subsequent works have explored diverse unlearning strategies, including clustered aggregation for asynchronous client removal \cite{ref-11}, differential privacy combined with gradient residual correction for certified unlearning \cite{ref-12}, auxiliary modules for simultaneous training and unlearning \cite{ref-13}, and orthogonal subspace descent to address gradient explosion \cite{ref-14}. Other contributions include class-discriminative pruning \cite{ref-15}, projected gradient ascent (PGA) \cite{ref-16}, and feature-level unlearning via Lipschitz sensitivity analysis \cite{ref-17}. Several comprehensive surveys \cite{ref-18,ref-19} have systematically categorized these methods and identified evaluation metrics to assess unlearning quality.

In spite of this progress, a critical assumption shared by nearly all existing FU methods remains largely unexamined: they are designed for the standard FedAvg paradigm \cite{ref-1}, where all clients collectively maintain a single global model. In practical deployments, however, statistical heterogeneity (non-IID distribution) of client data can severely degrade the performance of FedAvg \cite{ref-20}, which has led to the widespread adoption of personalized federated learning (pFL) \cite{ref-21}. pFL methods decompose the model into shared global components and client-specific personalized components, achieving substantially improved performance under data heterogeneity. Representative pFL approaches include FedPer \cite{ref-22}, which separates the model into a shared feature extractor and personalized classification heads; FedRep \cite{ref-23}, which learns shared representations while allowing clients to independently optimize local heads; Ditto \cite{ref-24}, which maintains personalized local models regularized toward the global model; and FedBN \cite{ref-25}, which personalizes batch normalization (BN) statistics while sharing all other parameters. These methods have demonstrated consistent improvements over FedAvg in heterogeneous settings \cite{ref-26}.

The architectural decomposition introduced by pFL fundamentally alters the semantics of FU, creating challenges that existing methods are not equipped to handle. Specifically, when a target client requests removal from the federation, the unlearning process must operate on two structurally distinct parameter spaces: the shared global layers, which aggregate contributions from all participating clients, and the personalized local layers, which encode each client's unique data distribution. This decomposition introduces a tension between two competing objectives. On the one hand, unlearning completeness requires thoroughly removing the target client's influence from the shared layers. On the other hand, aggressive modifications to these shared layers can disrupt the learned feature representations upon which remaining clients depend for their personalized performance, a phenomenon we term \textit{personalization degradation}. Existing FU methods, which treat all model parameters uniformly, are inherently unable to navigate this trade-off.

To date, only a few works have tangentially touched upon the intersection of FU and personalization. ZeroFU \cite{ref-27} proposed a zero-shot data-free unlearning approach using GAN-based distillation with personalized client features via conditional computation. Mimir \cite{ref-28} introduced a prompt-based personalized unlearning framework that uses client-specific prompts to capture data distribution characteristics. FUSED \cite{ref-29} analyzed layer-wise adapter structures for reversible unlearning. However, none of these methods systematically addresses unlearning under the model-splitting pFL paradigm (e.g., FedPer, FedRep) or regularization-based pFL (e.g., Ditto), nor do they provide formal analysis of the tension between shared and personalized layers.

Motivated by these observations, we propose pFedUL, a general layer-aware selective unlearning framework designed for pFL. The core insight is that different layer types require fundamentally different unlearning treatments: shared layers demand careful contribution removal with minimal collateral impact on remaining clients, while personalized layers of the target client can be directly discarded. pFedUL integrates Fisher information-based contribution attribution, adaptive selective correction, and lightweight local recalibration into a cohesive pipeline applicable across mainstream pFL architectures. We further introduce two pFL-specific evaluation metrics to assess unlearning quality dimensions not captured by existing benchmarks.
The main contributions of this paper are summarized as follows:
\begin{enumerate}[label=$\bullet$]
\item \textbf{Problem formulation:} We formalize the FU problem under the pFL paradigm, identifying a completeness-preservation trade-off between effective target-client removal on shared layers and personalization preservation for remaining clients.
\item \textbf{General framework:} We propose pFedUL, a layer-aware unlearning framework that combines Fisher-based layer-wise contribution attribution, adaptive selective correction with differentiated intensity across layer types, and a lightweight recalibration protocol for remaining clients, applicable across four mainstream pFL architectures (FedPer, FedRep, Ditto, and FedBN) with architecture-specific instantiations.
\item \textbf{Evaluation metrics:} We introduce two pFL-specific metrics---the Personalization Preservation Score (PPS) and the Cross-client Fairness Index (CFI)---to assess personalization preservation and cross-client fairness after unlearning, complementing existing MIA-based completeness measures.
\item \textbf{Empirical validation:} We conduct experiments on CIFAR-10, CIFAR-100, and FEMNIST across four pFL architectures, comparing pFedUL with six state-of-the-art FU methods adapted to the pFL setting and showing that pFedUL approaches retraining-level forgetting while better preserving personalization and substantially reducing unlearning cost.
\end{enumerate}

The remainder of this paper is organized as follows. Section 2 reviews related work on machine unlearning, FU, and pFL. Section 3 presents the proposed pFedUL framework, including the problem formulation and algorithmic details. Section 4 reports the experimental results and analysis. Section 5 discusses implications, limitations, and future directions. Finally, Section 6 concludes the paper.

\section{Related Work} \label{sect:s2}

\subsection{Machine Unlearning} \label{sect:s2dot1}

Machine unlearning aims to remove the influence of specific training data from a trained model without retraining from scratch \cite{ref-37}. Exact unlearning methods retrain the model on the remaining data after excluding target samples. SISA \cite{ref-6} improved efficiency by partitioning training data into disjoint shards, each associated with an independently trained sub-model, so that only the affected shard requires retraining upon a removal request. However, this sharding strategy is not directly applicable to distributed settings where data partitioning is determined by client boundaries.

Approximate unlearning methods seek to efficiently modify model parameters to approximate the effect of exact retraining. Influence functions \cite{ref-7} estimate parameter changes caused by removing a single data point through second-order Hessian-based approximations, though at prohibitive cost for large-scale models. Sekhari et al. \cite{ref-36} proposed certified unlearning algorithms with formal indistinguishability guarantees. Thudi et al. \cite{ref-38} analyzed gradient ascent as an unlearning mechanism and showed that simply reversing gradient updates does not guarantee complete forgetting under general conditions. Relatedly, Elastic Weight Consolidation (EWC) \cite{ref-40} uses the Fisher information matrix to quantify parameter importance for previously learned tasks, providing a foundational tool for identifying which parameters are most affected by specific data contributions.
Despite their effectiveness in centralized settings, these methods share a fundamental limitation: they assume direct access to training data or its statistics, which is unavailable in FL where data remain distributed across clients.

\subsection{FU} \label{sect:s2dot2}

FU extends machine unlearning to distributed settings, aiming to remove a target client's data contributions from the jointly trained global model without requiring full retraining or access to raw client data \cite{ref-9}. A diverse set of FU methods have been proposed over the past several years, which can be broadly categorized by their core technical approach. \tabref{tabref:table-fu} provides a comprehensive summary.

\begin{table}[htp]
\tablesize{\footnotesize}
\caption{Summary of existing FU methods. All methods assume the standard FedAvg paradigm or its direct variants. The rightmost column describes each method's consideration of pFL architectures.}
\label{tabref:table-fu}
\newcolumntype{C}{>{\centering\arraybackslash}X}
\newcolumntype{L}{>{\raggedright\arraybackslash}X}
\begin{tabularx}{\textwidth}{lllllL}
\toprule
\textbf{Method} & \textbf{Year} & \textbf{Venue} & \textbf{Technique} & \textbf{Granularity} & \textbf{Personalization Awareness}\\
\midrule
FedEraser \cite{ref-10} & 2021 & IWQoS & Gradient calibration & Client & Assumes a single global model shared by all clients\\
Rapid Retrain \cite{ref-30} & 2022 & INFOCOM & Taylor approximation & Sample & Operates on FedAvg; no personalized components considered\\
PGA \cite{ref-16} & 2022 & arXiv & PGA & Client & Uniform treatment of all parameters without layer distinction\\
Class-Pruning \cite{ref-15} & 2022 & WWW & Channel pruning & Class & Prunes global model channels; no client-specific layers involved\\
KNOT \cite{ref-11} & 2023 & INFOCOM & Clustered aggregation & Client & Clusters relate to grouped clients, not personalized architectures\\
MUFC \cite{ref-33} & 2023 & ICLR & Federated clustering & Client/Sample & Exact unlearning via cluster retraining; no pFL model structure\\
FedRecovery \cite{ref-12} & 2023 & IEEE TIFS & DP + residual correction & Client & Certified removal on a single global model\\
FRU \cite{ref-31} & 2023 & WSDM & Rollback + calibration & User & Designed for recommender systems under FedAvg\\
FFMU \cite{ref-35} & 2023 & ICML & Functional theory & Sample & Functional-space operations on a unified global model\\
SIFU \cite{ref-32} & 2024 & AISTATS & Sequential informed & Client & Provable bounds derived under FedAvg assumptions\\
FedAU \cite{ref-13} & 2024 & IJCAI & Auxiliary module & Client/Class & Auxiliary module tracks global contributions only\\
Ferrari \cite{ref-17} & 2024 & NeurIPS & Lipschitz sensitivity & Feature & Feature-level removal on a shared global model\\
VeriFi \cite{ref-19} & 2024 & IEEE TDSC & Verification framework & Client & Verifies unlearning on FedAvg; pFL verification unexplored\\
FedOSD \cite{ref-14} & 2025 & AAAI & Orthogonal subspace & Client & Orthogonal projection in global parameter space\\
NoT \cite{ref-34} & 2025 & CVPR & Weight negation & Client & Negates global weights; no shared/personalized distinction\\
ZeroFU \cite{ref-27} & 2025 & IJCAI & GAN distillation & Client & Embeds client-specific conditional features, but not evaluated on pFL architectures\\
Mimir \cite{ref-28} & 2025 & IEEE IoTJ & Prompt distillation & Client & Uses client-specific prompts for personalization, limited to prompt-tuning scenarios\\
FUSED \cite{ref-29} & 2025 & CVPR & Sparse adapter decomp. & Client & Analyzes layer-wise adapters, but does not formalize shared-personalized decomposition\\
\bottomrule
\end{tabularx}
\end{table}

\textbf{Gradient calibration and approximation.} FedEraser \cite{ref-10} stores historical gradient updates during training and calibrates the global model by excluding the target client's contributions. Liu et al. \cite{ref-30} proposed a rapid retraining approach based on first-order Taylor expansion. These methods are computationally efficient but require substantial storage for historical updates.

\textbf{Structural decomposition.} Another line of research exploits federation structure to accelerate unlearning. KNOT \cite{ref-11} organizes clients into clusters with independently maintained sub-models, so that removal affects only the relevant cluster. MUFC \cite{ref-33} demonstrated that federated clustering can achieve exact unlearning with significant speedups. However, these clustering-based approaches do not account for the heterogeneous model structures found in pFL.

\textbf{Direct parameter manipulation.} Several methods directly modify model parameters to erase a target client's influence. FedOSD \cite{ref-14} constrains unlearning updates within orthogonal subspaces for stable convergence. Halimi et al. \cite{ref-16} proposed PGA, which reverses the target client's influence while projecting updates to preserve utility, and NoT \cite{ref-34} negates specific layer weights to disrupt inter-layer co-adaptation. These methods treat all model parameters uniformly, which becomes problematic when the model contains structurally distinct shared and personalized components.

\textbf{Pruning, distillation, and specialized approaches.} Wang et al. \cite{ref-15} proposed class-discriminative pruning using TF-IDF scoring for class-level unlearning. ZeroFU \cite{ref-27} proposed a zero-shot approach using GAN-based distillation with client-specific conditional features, making it one of the few methods that partially accounts for client-specific characteristics. Che et al. \cite{ref-35} developed FFMU based on nonlinear functional theory, enabling simultaneous training and unlearning. Additional specialized methods include FedRecovery \cite{ref-12} for certified removal, FedAU \cite{ref-13} for continuous contribution tracking, Ferrari \cite{ref-17} for feature-level unlearning, VeriFi \cite{ref-19} for unlearning verification, FRU \cite{ref-31} for recommendation domains, and SIFU \cite{ref-32} for provable bounds.

\textbf{Gap analysis.} As summarized in \tabref{tabref:table-fu}, the vast majority of existing FU methods operate under the assumption of a single global model, without awareness of the shared-personalized parameter distinction that characterizes pFL architectures. Only three recent works partially incorporate personalized elements: ZeroFU \cite{ref-27} incorporates client-specific conditional features but does not evaluate under model-splitting pFL; Mimir \cite{ref-28} uses prompt-based personalization but is restricted to prompt-tuning scenarios; and FUSED \cite{ref-29} analyzes layer-wise adapter structures but does not formalize the shared-personalized layer distinction. None of these methods directly addresses the tension between unlearning completeness on shared layers and personalization preservation for remaining clients, which motivates the present work.

\subsection{pFL} \label{sect:s2dot3}

The standard FedAvg algorithm \cite{ref-1} trains a single global model by averaging locally updated parameters from participating clients. However, when client data distributions are highly heterogeneous (non-IID), this uniform aggregation leads to significant performance degradation \cite{ref-20,ref-44}. pFL addresses this challenge by allowing each client to maintain model components tailored to its local data characteristics \cite{ref-21}.

The most intuitive approach is to split the network architecture into shared and private components. FedPer \cite{ref-22} designates the lower layers as a shared feature extractor while keeping the upper classification layers local to each client. FedRep \cite{ref-23} adopts a similar split but introduces alternating optimization that decouples the update of shared representations from local heads. FedRoD \cite{ref-42} extends this line by maintaining both a generic and a personalized classifier head simultaneously.

Personalization can also be achieved through optimization-level mechanisms. Ditto \cite{ref-24} trains a global model via FedAvg and then optimizes a personalized local model for each client using proximal regularization that penalizes deviation from the global model. pFedMe \cite{ref-41} follows a similar philosophy but employs Moreau envelopes for stronger convergence guarantees. A more lightweight approach targets normalization layers: FedBN \cite{ref-25} keeps only BN layers local while sharing all other parameters, based on the observation that BN statistics encode domain-specific distributional information. More recently, FedALA \cite{ref-26} proposed learning element-wise adaptive aggregation weights for fine-grained personalization.

Despite their diversity, all pFL methods share a common structural property: the model parameters are decomposed into shared parameters $\theta^{s}$ that capture cross-client common knowledge, and personalized parameters $\theta^{p}$ that encode client-specific patterns. This decomposition is fundamental to their effectiveness but introduces a structural complexity that has been largely overlooked by the FU literature. When a client requests removal, the unlearning procedure must selectively operate on $\theta^{s}$ to erase the target client's contributions while preserving the integrity of the shared representation upon which all remaining clients' personalized parameters $\theta^{p}_{i}$ depend. To the best of our knowledge, no existing work has formally studied this problem or proposed a dedicated solution, which motivates the framework presented in this paper.

\section{The Proposed pFedUL Framework} \label{sect:s3}

\subsection{Problem Formulation} \label{sect:s3dot1}

Consider an FL system consisting of $N$ clients $\{c_1, c_2, \ldots, c_N\}$, where each client $c_i$ has a private local dataset $\mathcal{D}_i$. The standard FL objective is to minimize the global empirical risk:
\begin{equation}
\min_{\theta} \; F(\theta) = \sum_{i=1}^{N} \frac{|\mathcal{D}_i|}{|\mathcal{D}|} F_i(\theta), \quad F_i(\theta) = \frac{1}{|\mathcal{D}_i|} \sum_{(x,y) \in \mathcal{D}_i} \ell(\theta; x, y),
\label{eq:1}
\end{equation}
where $\theta$ denotes the model parameters, $\mathcal{D} = \bigcup_{i=1}^{N} \mathcal{D}_i$ is the union of all local datasets, $\ell$ is the loss function, and $F_i(\theta)$ is the local objective of client $c_i$.

In pFL, the model parameters are decomposed into two functionally distinct groups: shared parameters $\theta^s$ that are aggregated across clients, and personalized parameters $\theta^p_i$ that are maintained locally by each client $c_i$. The pFL objective becomes:
\begin{equation}
\min_{\theta^s, \{\theta^p_i\}_{i=1}^{N}} \; \sum_{i=1}^{N} \frac{|\mathcal{D}_i|}{|\mathcal{D}|} F_i(\theta^s, \theta^p_i).
\label{eq:2}
\end{equation}
This general formulation encompasses the four mainstream pFL architectures considered in this work. In FedPer \cite{ref-22} and FedRep \cite{ref-23}, $\theta^s$ corresponds to the feature extractor layers and $\theta^p_i$ to the classification head of client $c_i$. In Ditto \cite{ref-24}, $\theta^s$ is the global model obtained through FedAvg, and $\theta^p_i$ is the local fine-tuned model regularized toward $\theta^s$ via a proximal term $\lambda \|\theta^p_i - \theta^s\|^2$. In FedBN \cite{ref-25}, $\theta^s$ consists of all parameters except BN layers, while $\theta^p_i$ contains the local BN statistics.

To clarify the architecture-specific differences in how shared and personalized parameters are defined, \tabref{tabref:table-arch} provides a detailed comparison across the four pFL methods. As shown, the nature and scale of personalized parameters vary substantially: in FedPer and FedRep, $\theta^p_i$ consists of a lightweight classification head (approximately 0.5\% of total parameters for ResNet-18 on CIFAR-10); in FedBN, $\theta^p_i$ comprises the BN layers (approximately 1.2\% of total parameters); while in Ditto, $\theta^p_i$ is a full copy of the entire model. These differences have direct implications for pFedUL's operation: the recalibration cost in Stage~3 is proportional to $|\theta^p_i|$, and the treatment of the target client's personalized parameters upon removal differs accordingly (discarding a classification head versus discarding a full model copy).

\begin{table}[ht]
\tablesize{\footnotesize}
\caption{Architecture-specific decomposition of shared parameters $\theta^s$ and personalized parameters $\theta^p_i$ across the four pFL methods evaluated in this work. The table specifies the concrete parameter assignment, the approximate proportion of personalized parameters relative to the total model, and the pFedUL-specific treatment during unlearning and recalibration for each architecture.}
\label{tabref:table-arch}
\newcolumntype{C}{>{\centering\arraybackslash}X}
\newcolumntype{L}{>{\raggedright\arraybackslash}X}
\begin{tabularx}{\textwidth}{lLLcLL}
\toprule
\textbf{Method} & \textbf{$\theta^s$ (Shared)} & \textbf{$\theta^p_i$ (Personalized)} & \textbf{$|\theta^p_i|/|\theta|$} & \textbf{Target $\theta^p_{c_t}$ Treatment} & \textbf{Recalibration} \\
\midrule
FedPer & All layers except final FC & Final FC layer (classifier head) & $\sim$0.5\% & Delete head & Fine-tune head with $\hat{\theta}^s$ frozen \\
FedRep & All layers except final FC & Final FC layer (classifier head) & $\sim$0.5\% & Delete head & Alternating opt. on head with $\hat{\theta}^s$ frozen \\
Ditto & Global model (full FedAvg model) & Full local model ($\ell_2$-regularized toward $\theta^s$) & 100\% & Delete full local model & Re-run proximal opt. (Eq.~\eqref{eq:13}) with $\hat{\theta}^s$ as anchor \\
FedBN & All layers except BN layers & BN running mean and variance & $\sim$1.2\% & Delete local BN stats & Re-estimate BN stats via forward pass on local data \\
\bottomrule
\end{tabularx}
\end{table}

We now formalize the FU problem under the pFL paradigm. Let $c_t$ denote the target client that requests data removal. After the pFL training process converges, the system maintains the shared parameters $\theta^s$ and each remaining client $c_i$ ($i \neq t$) maintains personalized parameters $\theta^p_i$. The objective of pFL-aware FU is to produce updated shared parameters $\hat{\theta}^s$ such that:
\begin{equation}
\hat{\theta}^s \approx \theta^{s,*}_{\setminus t} = \arg\min_{\theta^s} \sum_{i \neq t} \frac{|\mathcal{D}_i|}{|\mathcal{D}_{\setminus t}|} F_i(\theta^s, \theta^p_i),
\label{eq:3}
\end{equation}
where $\theta^{s,*}_{\setminus t}$ represents the shared parameters that would be obtained by retraining from scratch on the remaining clients, and $\mathcal{D}_{\setminus t} = \bigcup_{i \neq t} \mathcal{D}_i$.
This formulation reveals a fundamental tension unique to the pFL setting. The personalized parameters $\theta^p_i$ of the remaining clients were optimized jointly with the original shared parameters $\theta^s$, which contain the target client's contributions. Modifying $\theta^s$ to $\hat{\theta}^s$ for unlearning purposes disrupts this co-optimization, potentially degrading the personalized performance of remaining clients. We refer to this as the \textit{completeness-preservation trade-off}: aggressive unlearning on $\theta^s$ improves forgetting completeness but risks damaging remaining clients' personalization, while conservative unlearning preserves personalization but may leave residual traces of the target client's data.

\textbf{Remark 1} (Factors governing the trade-off severity). The severity of the completeness-preservation trade-off is governed by several factors that can be qualitatively characterized. First, when the target client's data volume $|\mathcal{D}_t|$ constitutes a larger fraction of the total data $|\mathcal{D}|$, its contributions to the shared parameters $\theta^s$ are more deeply embedded, requiring larger parameter modifications for complete removal and thus increasing the risk of disrupting remaining clients' personalization. Second, higher data heterogeneity (smaller $\alpha_{\text{dir}}$) leads to more differentiated per-client contributions across layers, which paradoxically benefits layer-aware unlearning by concentrating the target client's influence in fewer layers and thereby reducing collateral perturbation. Third, when remaining clients' personalized parameters $\theta^p_i$ have been tightly co-optimized with $\theta^s$ over many communication rounds, the sensitivity of $\theta^p_i$ to changes in $\theta^s$ increases, amplifying the misalignment caused by unlearning. These observations motivate the three-stage design of pFedUL, which uses Fisher-based attribution to identify high-influence layers (addressing the first factor), applies selective correction to exploit the concentration of contributions (addressing the second), and performs lightweight recalibration to restore co-optimization alignment (addressing the third). The empirical sensitivity analysis in Section~\ref{sect:s4dot5} provides quantitative support for these qualitative observations.

\subsection{Framework Overview} \label{sect:s3dot2}

\begin{figure}[ht]
\centering
\includegraphics[width=\textwidth]{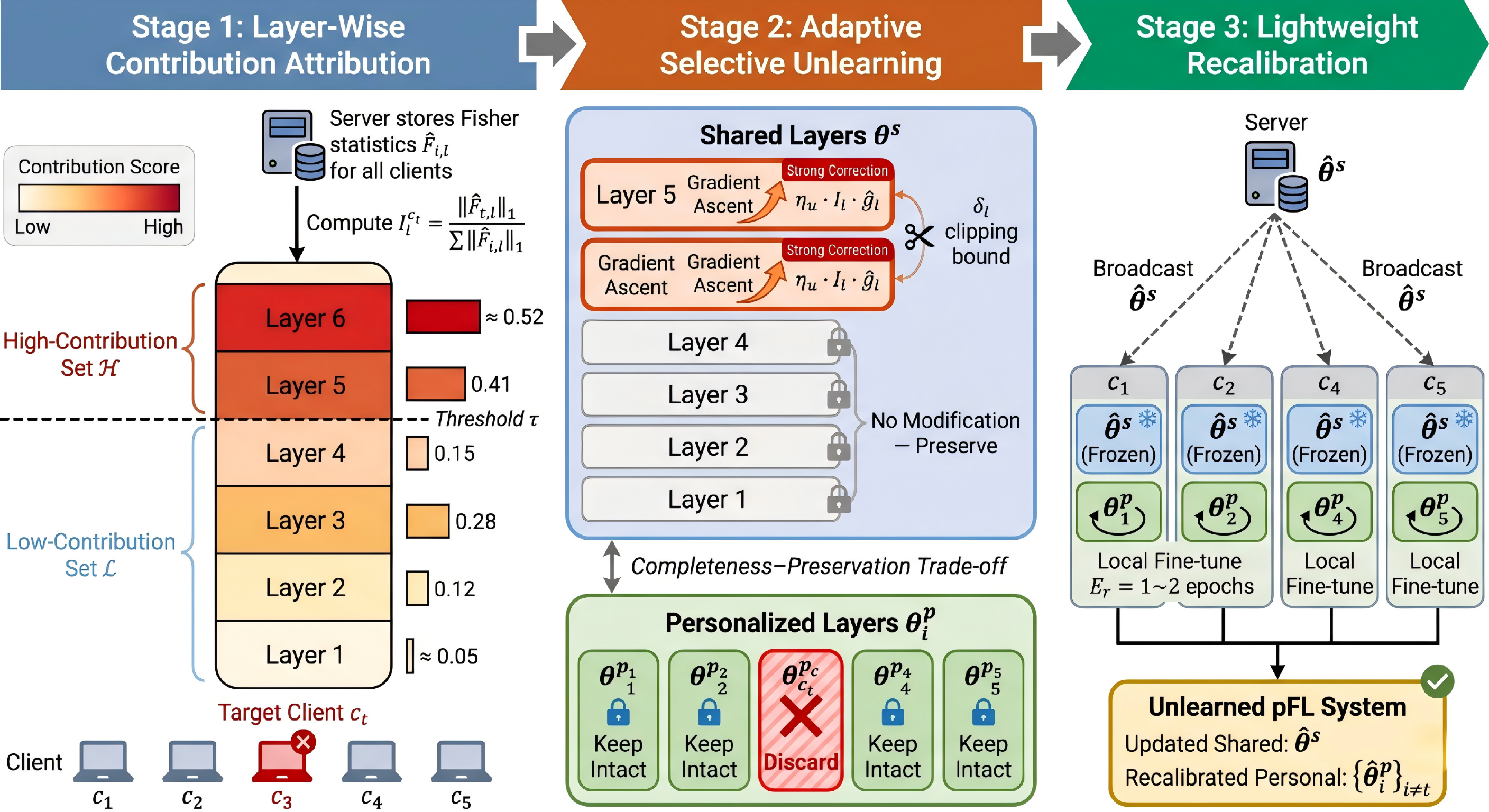}
\caption{Overview of the proposed pFedUL framework. The framework operates in three stages: (1) computing layer-wise contribution scores of the target client using Fisher-based attribution; (2) applying adaptive unlearning with differentiated intensity across shared layers based on contribution scores, while directly discarding the target client's personalized parameters; (3) performing lightweight local recalibration for remaining clients to restore personalized performance.}
\label{fig:framework}
\end{figure}

To address the completeness-preservation trade-off, pFedUL adopts a three-stage pipeline, as illustrated in \fig{fig:framework}. Given a trained pFL system and a target client $c_t$ requesting removal, pFedUL proceeds as follows.

In the first stage (Section~\ref{sect:s3dot3}), pFedUL computes a layer-wise contribution score for each shared layer, quantifying how much the target client $c_t$ has influenced each layer's parameters during training. This is achieved through a Fisher information-based attribution mechanism that operates on stored gradient statistics without requiring access to the target client's raw data.

In the second stage (Section~\ref{sect:s3dot4}), pFedUL applies adaptive selective unlearning to shared parameters $\theta^s$. Layers with high contribution scores from $c_t$ receive aggressive correction through a constrained gradient ascent, whereas layers with low contribution scores undergo minimal or no modification. The personalized parameters of the target client $\theta^p_{c_t}$ are immediately discarded, and all remaining personalized parameters of the clients $\theta^p_i$ ($i \neq t$) are kept intact during this stage.

In the third stage (Section~\ref{sect:s3dot5}), a lightweight recalibration protocol allows each remaining client to local fine-tune its personalized parameters $\theta^p_i$ with respect to the updated shared parameters $\hat{\theta}^s$, using only 1 to 2 epochs of local training. This restores the co-optimization relationship between shared and personalized components without requiring additional global communication rounds.

\subsection{Layer-Wise Contribution Attribution} \label{sect:s3dot3}

The first stage of pFedUL aims to quantify the target client's influence on each layer of the shared parameters. We propose a Fisher information-based attribution mechanism that provides a principled measure of each layer's sensitivity to the target client's data.

Let $\theta^s = \{\theta^s_1, \theta^s_2, \ldots, \theta^s_L\}$ denote the shared parameters decomposed into $L$ layers. During the federated training process, we maintain running estimates of the diagonal Fisher information matrix for each client and each layer. The diagonal Fisher information captures the expected squared gradient of the loss with respect to each parameter, serving as a measure of how sensitive a given layer is to the data of a particular client. Specifically, for client $c_i$ and layer $l$, the diagonal Fisher information is defined as:
\begin{equation}
\mathbf{F}_{i,l} = \frac{1}{|\mathcal{D}_i|} \sum_{(x,y) \in \mathcal{D}_i} \left( \nabla_{\theta^s_l} \ell(\theta^s, \theta^p_i; x, y) \right)^2.
\label{eq:4}
\end{equation}

In practice, computing Eq.~\eqref{eq:4} exactly requires access to client data, which is unavailable at unlearning time. Instead, we accumulate an exponential moving average (EMA) of the squared gradients during training. In each communication round $r$, when client $c_i$ uploads its local gradient update $g^r_{i,l}$ for layer $l$, the server maintains:
\begin{equation}
\hat{\mathbf{F}}_{i,l} \leftarrow \beta \hat{\mathbf{F}}_{i,l} + (1 - \beta) (g^r_{i,l})^2,
\label{eq:5}
\end{equation}
where $\beta \in (0,1)$ is the momentum coefficient that controls the trade-off between the recent and historical gradient information. Over the course of training, this running estimate converges to an approximation of the diagonal Fisher information.

Given the accumulated Fisher statistics, the layer-wise contribution score of the target client $c_t$ for layer $l$ is defined as the ratio of the target client's Fisher information to the aggregate Fisher information across all clients:
\begin{equation}
I^{c_t}_l = \frac{\|\hat{\mathbf{F}}_{t,l}\|_1}{\sum_{i=1}^{N} \|\hat{\mathbf{F}}_{i,l}\|_1},
\label{eq:6}
\end{equation}
where $\|\cdot\|_1$ denotes the $L_1$ norm. Intuitively, $I^{c_t}_l \in [0, 1]$ measures the proportion of the total parameter sensitivity at layer $l$ that is attributable to the target client. A high value of $I^{c_t}_l$ indicates that the target client has strongly shaped the parameters of layer $l$, suggesting that more aggressive unlearning is warranted at this layer.

This attribution mechanism offers two practical advantages. First, it only requires storing $L$ vectors of the same dimensionality as each layer's parameters per client, which incurs modest memory overhead. Second, the running average in Eq.~\eqref{eq:5} can be computed incrementally during training without requiring any post-hoc data access, making it naturally compatible with the privacy constraints of FL.

\textbf{Remark 2} (Theoretical connection between Fisher attribution and unlearning). The use of the diagonal Fisher information matrix as a contribution attribution metric is grounded in its well-established role as a measure of parameter importance with respect to specific data \cite{ref-40,ref-7}. For a given client $c_i$ and layer $l$, the diagonal Fisher $\mathbf{F}_{i,l}$ captures the expected curvature of the loss landscape along each parameter dimension: parameters with large Fisher values are those whose perturbation would most strongly affect the loss evaluated on client $c_i$'s data, and hence are the parameters that most encode $c_i$'s data characteristics. Consequently, the contribution score $I^{c_t}_l$ (Eq.~\eqref{eq:6}) quantifies the relative proportion of parameter sensitivity at layer $l$ that is attributable to the target client $c_t$. A high $I^{c_t}_l$ value indicates that the parameters of layer $l$ are disproportionately sensitive to the target client's data distribution, implying that the target client's influence is concentrated in that layer. By directing unlearning corrections to layers with high $I^{c_t}_l$, pFedUL targets the parameter subspace where the target client's data patterns are most strongly encoded, while preserving layers where the target client's influence is negligible relative to other clients. This reasoning is consistent with the EWC framework \cite{ref-40,ref-39}, which demonstrates that the Fisher information matrix reliably identifies parameters critical to specific tasks, and with influence function theory \cite{ref-7}, which shows that parameter importance weighted by curvature provides accurate approximations of data removal effects. We provide empirical validation of this theoretical connection in Section~\ref{sect:s4dot4b}, where we demonstrate that selectively unlearning high-attribution layers achieves substantially better forgetting than unlearning low-attribution layers.

\subsection{Adaptive Selective Unlearning} \label{sect:s3dot4}

Given the layer-wise contribution scores $\{I^{c_t}_l\}_{l=1}^{L}$ computed in the attribution stage, pFedUL applies differentiated unlearning strategies to different layers of the shared parameters.

We first classify each shared layer into one of two categories based on an adaptive threshold $\tau$. Layers with contribution scores exceeding $\tau$ are designated as \textit{high-contribution layers}, while the remaining layers are designated as \textit{low-contribution layers}:
\begin{equation}
\mathcal{H} = \{l \mid I^{c_t}_l > \tau\}, \quad \mathcal{L} = \{l \mid I^{c_t}_l \leq \tau\}.
\label{eq:7}
\end{equation}

The threshold $\tau$ is set adaptively based on the distribution of contribution scores:
\begin{equation}
\tau = \mu_I + \alpha \cdot \sigma_I,
\label{eq:8}
\end{equation}
where $\mu_I = \frac{1}{L}\sum_{l=1}^{L} I^{c_t}_l$ and $\sigma_I = \sqrt{\frac{1}{L}\sum_{l=1}^{L}(I^{c_t}_l - \mu_I)^2}$ are the mean and standard deviation of the contribution scores across all layers, and $\alpha \geq 0$ is a sensitivity hyperparameter. A larger $\alpha$ yields a more conservative unlearning strategy that modifies only the most heavily influenced layers, while $\alpha = 0$ applies correction to all above-average layers.

For high-contribution layers ($l \in \mathcal{H}$), we apply constrained gradient ascent to actively erase the target client's influence. The update for layer $l$ is:
\begin{equation}
\hat{\theta}^s_l = \theta^s_l + \eta_u \cdot I^{c_t}_l \cdot \hat{g}^{c_t}_l,
\label{eq:9}
\end{equation}
where $\eta_u$ is the unlearning step size and $\hat{g}^{c_t}_l$ is the estimated unlearning direction for layer $l$. The unlearning direction is derived from the accumulated gradient information of the target client. Specifically, it combines the Fisher sensitivity of each parameter with its deviation from initialization, under the heuristic that parameters exhibiting both high sensitivity to the target client and large displacement from their initial values are most likely to encode that client's data patterns:
\begin{equation}
\hat{g}^{c_t}_l = \hat{\mathbf{F}}_{t,l} \odot (\theta^s_l - \theta^s_{l,\text{init}}),
\label{eq:10}
\end{equation}
where $\theta^s_{l,\text{init}}$ denotes the initial parameters of layer $l$ before training, and $\odot$ denotes element-wise multiplication.

The multiplication by $I^{c_t}_l$ in Eq.~\eqref{eq:9} ensures that the correction intensity is proportional to the layer's contribution score, implementing the adaptive nature of the approach. To prevent the unlearning update from excessively degrading model utility, we apply a projection constraint that bounds the magnitude of the modification at each layer:
\begin{equation}
\hat{\theta}^s_l \leftarrow \theta^s_l + \text{clip}\left(\hat{\theta}^s_l - \theta^s_l, \; -\delta_l, \; \delta_l\right),
\label{eq:11}
\end{equation}
where $\delta_l = \gamma \cdot \|\theta^s_l\|_2 / \sqrt{d_l}$ is the per-layer clipping bound, $d_l$ is the dimensionality of layer $l$, and $\gamma$ is a global clipping coefficient. This projection ensures that the unlearning update remains within a trust region around the original parameters, reducing the risk of catastrophic degradation of the shared representation.

For low-contribution layers ($l \in \mathcal{L}$), no modification is applied, i.e., $\hat{\theta}^s_l = \theta^s_l$. This selective approach minimizes unnecessary perturbation to the shared representation, which is the key mechanism for preserving remaining clients' personalization.

Regarding personalized parameters, the target client's personalized parameters $\theta^p_{c_t}$ are directly deleted from the system, as they are exclusively owned by the departing client. The personalized parameters $\theta^p_i$ of all remaining clients ($i \neq t$) are kept unchanged during this stage. Any potential misalignment with the updated $\hat{\theta}^s$ is addressed in the subsequent recalibration stage.

\textit{Remark.} The layer-aware strategy is expected to outperform uniform unlearning (applying the same correction to all layers) for two reasons. First, in deep neural networks, the influence of a specific client's data is typically concentrated in certain layers rather than uniformly distributed \cite{ref-40}. Second, modifying low-contribution layers introduces unnecessary noise that degrades the shared representation without improving forgetting completeness, whereas the selective approach avoids this wasteful perturbation. This reasoning is empirically validated in the ablation study (Section~\ref{sect:s4dot4}).

\subsection{Lightweight Recalibration Protocol} \label{sect:s3dot5}

After the selective unlearning stage modifies the shared parameters from $\theta^s$ to $\hat{\theta}^s$, the personalized parameters $\theta^p_i$ of remaining clients may no longer be optimally aligned with the updated shared representation. This misalignment arises because $\theta^p_i$ was jointly optimized with the original $\theta^s$, and the unlearning-induced modifications, even when carefully controlled, shift the feature space encoded by the shared layers.

To restore this alignment, pFedUL employs a lightweight recalibration protocol. Each remaining client $c_i$ ($i \neq t$) receives the updated shared parameters $\hat{\theta}^s$ and performs a small number of local training epochs $E_r$ to re-optimize its personalized parameters:
\begin{equation}
\hat{\theta}^p_i = \arg\min_{\theta^p_i} F_i(\hat{\theta}^s, \theta^p_i),
\label{eq:12}
\end{equation}
where $\hat{\theta}^s$ is frozen during this local optimization. In practice, we find that $E_r = 1$ to $2$ local epochs are sufficient to restore personalized performance to within approximately 1\% of the pre-unlearning level, as shown in our experiments.

The recalibration protocol is efficient for three reasons. First, only the personalized parameters $\theta^p_i$ are updated, which typically constitute a small fraction of the total model parameters (e.g., only the classification head in FedPer/FedRep, or only the BN layers in FedBN). Second, the shared parameters $\hat{\theta}^s$ are frozen, so no global communication is required during recalibration. Third, the number of recalibration epochs $E_r$ is very small because the selective unlearning strategy in Section~\ref{sect:s3dot4} has already minimized disruption to low-contribution layers, reducing the magnitude of misalignment that recalibration needs to correct.

For the special case of Ditto, where the personalized model $\theta^p_i$ is a full local model regularized toward $\theta^s$, the recalibration takes the form:
\begin{equation}
\hat{\theta}^p_i = \arg\min_{\theta^p_i} \left[ F_i(\theta^p_i) + \frac{\lambda}{2}\|\theta^p_i - \hat{\theta}^s\|^2 \right],
\label{eq:13}
\end{equation}
which re-runs the Ditto local optimization with the updated global model $\hat{\theta}^s$ as the new regularization anchor. This requires $E_r$ local epochs with no additional communication, maintaining the same overhead as in the model-splitting case.

\subsection{Evaluation Metrics for pFL Unlearning} \label{sect:s3dot6}

Standard FU evaluation relies on metrics such as membership inference attack (MIA) accuracy \cite{ref-39} for unlearning completeness and remaining accuracy for model utility preservation. While these metrics remain relevant, they do not capture two quality dimensions that are important in the pFL setting: the preservation of per-client personalized performance and the fairness of unlearning impact across remaining clients. We introduce two complementary metrics to address these gaps.

The PPS measures the extent to which each remaining client retains its personalized performance after unlearning. Let $A^{\text{pre}}_i$ and $A^{\text{post}}_i$ denote the personalized test accuracy of client $c_i$ before and after unlearning, respectively. The PPS is defined as:
\begin{equation}
\text{PPS} = \frac{1}{N-1} \sum_{i \neq t} \frac{A^{\text{post}}_i}{A^{\text{pre}}_i}.
\label{eq:14}
\end{equation}

A PPS of 1.0 indicates perfect personalization preservation, meaning that no remaining client has experienced any performance degradation due to unlearning. Values below 1.0 indicate degradation, with lower values signifying more severe impact. Note that the Retrain baseline is assigned PPS~$= 1.0$ and CFI~$= 1.0$ by definition, since retraining from scratch produces the ideal reference model; all other methods are evaluated relative to this upper bound. Unlike global remaining accuracy, PPS explicitly accounts for each client's individual performance trajectory, making it sensitive to cases where the overall average accuracy is maintained but specific clients suffer disproportionate losses.

The CFI quantifies the uniformity of unlearning impact across remaining clients. An ideal unlearning procedure should affect all remaining clients similarly, rather than disproportionately harming certain clients. The CFI is defined as:
\begin{equation}
\text{CFI} = 1 - \sqrt{\frac{1}{N-1} \sum_{i \neq t} \left( \frac{A^{\text{post}}_i}{A^{\text{pre}}_i} - \text{PPS} \right)^2}.
\label{eq:15}
\end{equation}
The CFI subtracts the standard deviation of per-client preservation ratios from 1, yielding values close to 1.0 when the unlearning impact is uniformly distributed and lower values when certain clients are disproportionately affected. Together, PPS and CFI provide a more comprehensive assessment of pFL unlearning quality that complements the standard MIA-based completeness metrics.

\textbf{Remark 3} (Limitations of PPS and CFI). While PPS and CFI capture important evaluation dimensions, several caveats should be noted. First, both metrics are ratio-based, which makes them sensitive to clients with very low initial accuracy $A^{\text{pre}}_i$: a small absolute change in accuracy can produce a disproportionately large ratio deviation. In our experiments, this issue is mitigated by the fact that pFL methods generally achieve reasonable personalized accuracy for all clients, but in extreme heterogeneity settings where some clients have very few samples, this sensitivity warrants caution. Second, clients with small local test sets may exhibit high variance in their accuracy estimates, which propagates into both PPS and CFI. To account for this, we report standard deviations across five independent runs, and in Section~\ref{sect:s4dot4b} we additionally present client-level performance distributions to complement the averaged metrics. Third, PPS treats all clients equally regardless of their data volume, which may not always align with practical deployment priorities where certain clients may be more important than others. Despite these limitations, PPS and CFI provide a substantially more informative evaluation than global remaining accuracy alone, and we consider them useful additions to the FU evaluation toolkit.

\subsection{Overall Algorithm} \label{sect:s3dot7}

The complete pFedUL procedure is summarized in Algorithm~1. The algorithm takes as input the trained pFL system (shared parameters, personalized parameters, and stored Fisher statistics) and the identity of the target client, and outputs the unlearned shared parameters along with recalibrated personalized parameters for all remaining clients.

\begin{algorithm}[H]
\caption{The pFedUL Framework}
\label{alg:pfedulframework}
\begin{algorithmic}[1]
\Require Shared parameters $\theta^s = \{\theta^s_1, \ldots, \theta^s_L\}$; personalized parameters $\{\theta^p_i\}_{i=1}^N$; stored Fisher statistics $\{\hat{\mathbf{F}}_{i,l}\}$ for all clients $i$ and layers $l$; target client $c_t$; hyperparameters $\alpha$, $\eta_u$, $\gamma$, $E_r$
\Ensure Unlearned shared parameters $\hat{\theta}^s$; recalibrated personalized parameters $\{\hat{\theta}^p_i\}_{i \neq t}$
\Statex
\Statex \textit{// Stage 1: Layer-Wise Contribution Attribution}
\For{each shared layer $l = 1, 2, \ldots, L$}
    \State Compute contribution score $I^{c_t}_l \leftarrow \|\hat{\mathbf{F}}_{t,l}\|_1 \;/\; \sum_{i=1}^{N} \|\hat{\mathbf{F}}_{i,l}\|_1$ \Comment{Eq.~\eqref{eq:6}}
\EndFor
\State Compute threshold $\tau \leftarrow \mu_I + \alpha \cdot \sigma_I$ \Comment{Eq.~\eqref{eq:8}}
\State Classify layers: $\mathcal{H} \leftarrow \{l \mid I^{c_t}_l > \tau\}$, $\mathcal{L} \leftarrow \{l \mid I^{c_t}_l \leq \tau\}$ \Comment{Eq.~\eqref{eq:7}}
\Statex
\Statex \textit{// Stage 2: Adaptive Selective Unlearning}
\For{each layer $l \in \mathcal{H}$}
    \State Compute unlearning direction $\hat{g}^{c_t}_l \leftarrow \hat{\mathbf{F}}_{t,l} \odot (\theta^s_l - \theta^s_{l,\text{init}})$ \Comment{Eq.~\eqref{eq:10}}
    \State Compute update $\Delta_l \leftarrow \eta_u \cdot I^{c_t}_l \cdot \hat{g}^{c_t}_l$ \Comment{Eq.~\eqref{eq:9}}
    \State Compute clipping bound $\delta_l \leftarrow \gamma \cdot \|\theta^s_l\|_2 / \sqrt{d_l}$
    \State Apply projection $\hat{\theta}^s_l \leftarrow \theta^s_l + \text{clip}(\Delta_l, -\delta_l, \delta_l)$ \Comment{Eq.~\eqref{eq:11}}
\EndFor
\For{each layer $l \in \mathcal{L}$}
    \State $\hat{\theta}^s_l \leftarrow \theta^s_l$ \Comment{No modification}
\EndFor
\State Delete target client's personalized parameters $\theta^p_{c_t}$
\Statex
\Statex \textit{// Stage 3: Lightweight Recalibration}
\State Broadcast $\hat{\theta}^s$ to all remaining clients
\For{each remaining client $c_i$ ($i \neq t$) \textbf{in parallel}}
    \For{epoch $= 1, 2, \ldots, E_r$}
        \State Update $\theta^p_i \leftarrow \theta^p_i - \eta_r \nabla_{\theta^p_i} F_i(\hat{\theta}^s, \theta^p_i)$ \Comment{Eq.~\eqref{eq:12}}
    \EndFor
    \State $\hat{\theta}^p_i \leftarrow \theta^p_i$
\EndFor
\State \Return $\hat{\theta}^s$, $\{\hat{\theta}^p_i\}_{i \neq t}$
\end{algorithmic}
\end{algorithm}

The computational complexity of Algorithm~1 is dominated by Stages~2 and~3. Stage~1 requires only $O(L)$ vector norm computations. Stage~2 performs element-wise operations on at most $|\mathcal{H}|$ layers, with cost $O(\sum_{l \in \mathcal{H}} d_l)$, which is linear in the number of parameters in the high-contribution layers. Stage~3 requires each remaining client to perform $E_r$ local epochs on their personalized parameters, with cost $O(E_r \cdot |\theta^p_i| \cdot |\mathcal{D}_i|)$ per client. Since $|\theta^p_i| \ll |\theta^s|$ in typical pFL architectures and $E_r$ is very small (1 to 2), the total overhead of pFedUL is substantially lower than retraining from scratch, which would require $O(T \cdot |\theta| \cdot |\mathcal{D}_{\setminus t}|)$ for $T$ communication rounds over the full parameter set.

\section{Experiments} \label{sect:s4}

\subsection{Experimental Setup} \label{sect:s4dot1}

We evaluate pFedUL on three widely used FL benchmarks. CIFAR-10 consists of 60,000 color images across 10 classes, with 50,000 for training and 10,000 for testing. CIFAR-100 follows the same image dimensions but spans 100 fine-grained classes, each containing 600 images. FEMNIST is a character recognition dataset from the LEAF benchmark \cite{ref-45} with naturally heterogeneous data distributions across writers, containing 62 classes (digits, uppercase, and lowercase letters). To simulate controlled non-IID data heterogeneity, we partition CIFAR-10 and CIFAR-100 across clients using the Dirichlet distribution $\text{Dir}(\alpha_{\text{dir}})$ \cite{ref-43}, where smaller $\alpha_{\text{dir}}$ values indicate higher heterogeneity. Unless otherwise stated, we set $\alpha_{\text{dir}} = 0.5$ as the default non-IID setting. For FEMNIST, we use the natural writer-based partition provided by the benchmark.

We deploy $N = 20$ clients in all experiments and designate one randomly selected client as the target client $c_t$ for unlearning. All models use ResNet-18 as the backbone architecture. The federated training runs for $T = 200$ communication rounds with full client participation, using SGD with learning rate 0.01, momentum 0.9, and local epochs $E = 5$. For pFedUL, the key hyperparameters are set as follows: Fisher momentum $\beta = 0.99$, threshold sensitivity $\alpha = 0.5$, unlearning step size $\eta_u = 0.1$, clipping coefficient $\gamma = 0.01$, and recalibration epochs $E_r = 2$. The selection of these hyperparameters is guided by the following considerations. The Fisher momentum $\beta = 0.99$ follows the standard EMA practice used in Adam-style optimizers, ensuring that the accumulated Fisher statistics reflect a smoothed estimate over the full training trajectory. The threshold sensitivity $\alpha = 0.5$ is selected based on the trade-off analysis presented in Section~\ref{sect:s4dot5}, which shows that $\alpha = 0.5$ provides a favorable balance between unlearning completeness and personalization preservation. The unlearning step size $\eta_u = 0.1$ is set to be 10$\times$ the training learning rate (0.01), consistent with the common practice that unlearning typically requires larger step sizes than learning to effectively reverse parameter updates within a single pass \cite{ref-38}. The clipping coefficient $\gamma = 0.01$ constrains per-layer modifications to approximately 1\% of the layer's parameter norm, which we found sufficient to prevent catastrophic degradation while allowing meaningful correction. The recalibration epoch count $E_r = 2$ is chosen because our experiments show diminishing returns beyond this point: increasing $E_r$ from 1 to 2 recovers the majority of misalignment-induced PPS loss (from 0.948 to 0.975), while further increasing $E_r$ to 5 yields negligible additional improvement (0.977). All experiments are repeated five times with different random seeds, and we report the mean and standard deviation.

We evaluate pFedUL under four mainstream pFL architectures that span the major personalization paradigms. FedPer \cite{ref-22} splits the model at the last fully connected layer, sharing the feature extractor and keeping the classifier head local. FedRep \cite{ref-23} uses the same split but employs alternating optimization between shared and personalized components. Ditto \cite{ref-24} trains personalized local models with $\ell_2$ regularization ($\lambda = 0.1$) toward the global model. FedBN \cite{ref-25} keeps BN layers local while sharing all other parameters.

We compare pFedUL against five basic baselines and six state-of-the-art FU methods. The basic baselines are as follows. \textit{Retrain} removes the target client and retrains the entire pFL system from scratch, serving as the gold-standard upper bound. \textit{FedEraser-adapted} applies the FedEraser \cite{ref-10} gradient calibration method to the shared parameters $\theta^s$ of each pFL architecture. \textit{GA-Naive} performs uniform gradient ascent on all shared layers without layer-wise differentiation. \textit{Fine-tune} first applies gradient ascent for unlearning and then fine-tunes the entire model on remaining clients' data for 10 rounds. \textit{No-Unlearn} makes no modifications after the target client departs, serving as a lower bound for unlearning completeness.

To provide a more comprehensive comparison, we further adapt six representative FU methods to the pFL setting by applying their unlearning operations to the shared parameters $\theta^s$ while preserving the personalized parameters of remaining clients. These include: \textit{KNOT-adapted} \cite{ref-11}, which performs clustered aggregation-based unlearning on shared parameters; \textit{FedRecovery-adapted} \cite{ref-12}, which combines differential privacy with gradient residual correction on shared layers; \textit{PGA-adapted} \cite{ref-16}, which applies PGA on shared layers; \textit{ZeroFU-adapted} \cite{ref-27}, which uses GAN-based distillation with client-specific conditional features; \textit{Mimir-adapted} \cite{ref-28}, which employs prompt-based knowledge distillation for personalized unlearning; and \textit{FUSED-adapted} \cite{ref-29}, which applies sparse adapter decomposition for layer-wise unlearning. For methods that incorporate personalization-aware components (ZeroFU, Mimir, FUSED), we use their original designs where applicable and adapt only the incompatible components to the pFL architectures under evaluation.

To ensure fair comparison, \tabref{tabref:table-adapt} details the specific adaptation protocol applied to each baseline method, including which parameters are modified during unlearning, which are frozen, how the target client's personalized parameters are handled, and whether a recalibration step is included. For methods that do not originally include any form of post-unlearning recalibration, we do not add one, as doing so would conflate the baseline's intrinsic capability with the benefit of recalibration. This design ensures that the observed improvements of pFedUL are attributable to its complete framework rather than an unfair advantage from recalibration alone. The ablation study in Section~\ref{sect:s4dot4} separately quantifies the contribution of recalibration.

\begin{table}[ht]
\tablesize{\footnotesize}
\caption{Adaptation protocol for baseline FU methods under the pFL setting. For each method, the table specifies which parameters are subject to unlearning operations, which are frozen, how the target client's personalized parameters are treated, and whether post-unlearning recalibration is applied. All adapted methods delete the target client's personalized parameters $\theta^p_{c_t}$ and preserve remaining clients' personalized parameters $\theta^p_i$ ($i \neq t$) unless otherwise noted.}
\label{tabref:table-adapt}
\newcolumntype{C}{>{\centering\arraybackslash}X}
\newcolumntype{L}{>{\raggedright\arraybackslash}X}
\begin{tabularx}{\textwidth}{lLcL}
\toprule
\textbf{Method} & \textbf{Parameters Modified} & \textbf{Recalib.} & \textbf{Adaptation Notes} \\
\midrule
GA-Naive & All shared layers $\theta^s$ (uniform GA) & No & Gradient ascent applied uniformly to all shared parameters \\
Fine-tune & All shared layers $\theta^s$ (GA then FT) & No & 10 rounds of fine-tuning on remaining clients after GA \\
FedEraser-adapt. & Shared layers $\theta^s$ (gradient calibration) & No & Historical gradients used to calibrate $\theta^s$ only \\
PGA-adapt. & Shared layers $\theta^s$ (projected GA) & No & Projection constraint applied in $\theta^s$ space \\
KNOT-adapt. & Shared layers $\theta^s$ (cluster retrain) & No & Cluster formed on shared parameters; affected cluster retrained \\
FedRecovery-adapt. & Shared layers $\theta^s$ (DP + residual) & No & DP noise and gradient residual correction on $\theta^s$ \\
ZeroFU-adapt.$^\dagger$ & Shared layers $\theta^s$ (GAN distillation) & No & Client-specific conditional features retained from original design \\
Mimir-adapt.$^\dagger$ & Shared layers $\theta^s$ (prompt distillation) & No & Client-specific prompts retained; distillation on $\theta^s$ \\
FUSED-adapt.$^\dagger$ & Shared layers $\theta^s$ (sparse adapters) & No & Layer-wise adapter decomposition applied to $\theta^s$ \\
pFedUL & High-contribution shared layers (selective) & Yes & Fisher-based selective correction + $E_r=2$ recalibration \\
\bottomrule
\end{tabularx}
\footnotesize{$^\dagger$Methods with partial personalization awareness in their original designs.}
\end{table}

We evaluate all methods using four metrics: MIA accuracy \cite{ref-39} (closer to 0.5 indicates better unlearning), Remaining Accuracy (test accuracy of remaining clients after unlearning), PPS (Eq.~\eqref{eq:14}), and CFI (Eq.~\eqref{eq:15}). Additionally, we measure the wall-clock time of each unlearning method.

\subsection{Unlearning Effectiveness} \label{sect:s4dot2}

We first evaluate unlearning completeness by measuring MIA accuracy and remaining model utility. \tabref{tabref:table-main} presents the results across all datasets, pFL architectures, and methods.

\begin{table}[H]
\tablesize{\scriptsize}
\caption{Unlearning effectiveness comparison across three datasets and four pFL architectures. MIA Acc. ($\downarrow$, closer to 0.50 is better) measures unlearning completeness; Rem. Acc. ($\uparrow$) measures remaining model utility. Best results (excluding Retrain) are in \textbf{bold}. Results are averaged over 5 runs.}
\label{tabref:table-main}
\newcolumntype{C}{>{\centering\arraybackslash}X}
\begin{tabularx}{\textwidth}{l *{8}{C}}
\toprule
& \multicolumn{2}{c}{\textbf{FedPer}} & \multicolumn{2}{c}{\textbf{FedRep}} & \multicolumn{2}{c}{\textbf{Ditto}} & \multicolumn{2}{c}{\textbf{FedBN}} \\
\cmidrule(lr){2-3} \cmidrule(lr){4-5} \cmidrule(lr){6-7} \cmidrule(lr){8-9}
\textbf{Method} & MIA$\downarrow$ & Rem.$\uparrow$ & MIA$\downarrow$ & Rem.$\uparrow$ & MIA$\downarrow$ & Rem.$\uparrow$ & MIA$\downarrow$ & Rem.$\uparrow$ \\
\midrule
\multicolumn{9}{l}{\textit{CIFAR-10 ($\alpha_{\text{dir}}=0.5$)}} \\
\midrule
No-Unlearn & 78.4 & 84.7 & 79.1 & 85.3 & 77.6 & 86.1 & 76.8 & 80.5 \\
Fine-tune & 58.3 & 82.1 & 59.1 & 83.0 & 57.5 & 84.3 & 58.7 & 78.6 \\
GA-Naive & 50.2 & 72.4 & 49.8 & 73.1 & 50.6 & 74.5 & 49.5 & 69.8 \\
FedEraser-adapt. & 53.6 & 78.9 & 54.1 & 79.5 & 53.2 & 80.8 & 54.8 & 75.3 \\
\textbf{pFedUL} & \textbf{51.3} & \textbf{83.5} & \textbf{51.1} & \textbf{84.2} & \textbf{51.5} & \textbf{85.4} & \textbf{51.8} & \textbf{79.6} \\
Retrain & 50.1 & 84.3 & 50.2 & 84.9 & 50.3 & 85.8 & 50.0 & 80.1 \\
\midrule
\multicolumn{9}{l}{\textit{CIFAR-100 ($\alpha_{\text{dir}}=0.5$)}} \\
\midrule
No-Unlearn & 81.2 & 58.6 & 82.0 & 59.4 & 80.3 & 60.7 & 79.5 & 55.2 \\
Fine-tune & 60.7 & 56.1 & 61.3 & 56.8 & 59.8 & 58.2 & 61.5 & 52.7 \\
GA-Naive & 51.4 & 44.3 & 50.8 & 45.1 & 51.9 & 46.8 & 50.3 & 41.6 \\
FedEraser-adapt. & 55.9 & 51.8 & 56.4 & 52.5 & 55.1 & 53.9 & 57.2 & 48.3 \\
\textbf{pFedUL} & \textbf{52.1} & \textbf{57.5} & \textbf{51.8} & \textbf{58.3} & \textbf{52.4} & \textbf{59.8} & \textbf{52.7} & \textbf{54.1} \\
Retrain & 50.3 & 58.1 & 50.1 & 59.0 & 50.5 & 60.3 & 50.2 & 54.8 \\
\midrule
\multicolumn{9}{l}{\textit{FEMNIST (natural partition)}} \\
\midrule
No-Unlearn & 75.6 & 91.2 & 76.3 & 91.8 & 74.9 & 92.4 & 74.1 & 89.3 \\
Fine-tune & 56.4 & 89.5 & 57.2 & 90.1 & 55.8 & 90.9 & 57.1 & 87.5 \\
GA-Naive & 50.8 & 82.6 & 50.3 & 83.2 & 51.2 & 84.1 & 49.7 & 80.4 \\
FedEraser-adapt. & 53.1 & 86.3 & 53.8 & 86.9 & 52.7 & 87.8 & 54.3 & 83.7 \\
\textbf{pFedUL} & \textbf{51.0} & \textbf{90.4} & \textbf{50.8} & \textbf{91.0} & \textbf{51.3} & \textbf{91.7} & \textbf{51.6} & \textbf{88.5} \\
Retrain & 50.2 & 90.8 & 50.0 & 91.5 & 50.4 & 92.1 & 50.1 & 89.0 \\
\bottomrule
\end{tabularx}
\end{table}

Several observations can be drawn from \tabref{tabref:table-main}. First, pFedUL consistently achieves MIA accuracy within 1.0 to 1.8 percentage points of the Retrain baseline across all dataset--architecture combinations, suggesting that the layer-aware unlearning strategy effectively erases the target client's data influence. In contrast, Fine-tune and FedEraser-adapted leave substantially higher MIA accuracy (7 to 11 points above 0.5), indicating incomplete forgetting. GA-Naive achieves MIA accuracy close to 0.5 but at the cost of severe remaining accuracy degradation, as the uniform gradient ascent indiscriminately damages all shared layers regardless of their contribution from the target client.

Second, pFedUL maintains remaining accuracy within 0.5 to 1.2 percentage points of Retrain, whereas GA-Naive suffers 10 to 15 point drops and FedEraser-adapted drops 4 to 7 points. This suggests that the selective nature of pFedUL's layer-aware correction successfully preserves shared representation quality. The advantage is particularly notable on CIFAR-100, where the larger number of classes makes the shared representation more sensitive to indiscriminate perturbation.

Third, the results are consistent across all four pFL architectures, supporting the generalizability of pFedUL's unified formulation. The slight variations across architectures (e.g., FedBN showing marginally lower remaining accuracy) likely reflect inherent differences in personalization capacity rather than limitations of pFedUL.

\subsection{Personalization Preservation} \label{sect:s4dot3}

While \tabref{tabref:table-main} shows pFedUL's effectiveness in balancing unlearning completeness and remaining accuracy, the PPS and CFI metrics provide deeper insight into per-client personalization quality. \tabref{tabref:table-pps} presents these results with standard deviations across five runs. Note that Retrain serves as the normalized upper bound (PPS~$= 1.0$, CFI~$= 1.0$), and all other methods are evaluated against this reference.

\begin{table}[H]
\tablesize{\scriptsize}
\caption{PPS ($\uparrow$) and CFI ($\uparrow$) across datasets and pFL architectures. Mean $\pm$ std over 5 runs. Best results (excluding Retrain) are in \textbf{bold}. $\Delta$ denotes the improvement of pFedUL over the second-best baseline.}
\label{tabref:table-pps}
\newcolumntype{C}{>{\centering\arraybackslash}X}
\begin{tabularx}{\textwidth}{l *{8}{C}}
\toprule
& \multicolumn{2}{c}{\textbf{FedPer}} & \multicolumn{2}{c}{\textbf{FedRep}} & \multicolumn{2}{c}{\textbf{Ditto}} & \multicolumn{2}{c}{\textbf{FedBN}} \\
\cmidrule(lr){2-3} \cmidrule(lr){4-5} \cmidrule(lr){6-7} \cmidrule(lr){8-9}
\textbf{Method} & PPS$\uparrow$ & CFI$\uparrow$ & PPS$\uparrow$ & CFI$\uparrow$ & PPS$\uparrow$ & CFI$\uparrow$ & PPS$\uparrow$ & CFI$\uparrow$ \\
\midrule
\multicolumn{9}{l}{\textit{CIFAR-10 ($\alpha_{\text{dir}}=0.5$)}} \\
\midrule
Fine-tune & .941$\pm$.012 & .903$\pm$.018 & .946$\pm$.010 & .911$\pm$.015 & .952$\pm$.009 & .918$\pm$.014 & .934$\pm$.014 & .895$\pm$.021 \\
GA-Naive & .827$\pm$.021 & .784$\pm$.032 & .833$\pm$.019 & .791$\pm$.028 & .841$\pm$.018 & .802$\pm$.025 & .819$\pm$.024 & .771$\pm$.035 \\
FedEraser-adapt. & .862$\pm$.016 & .831$\pm$.023 & .869$\pm$.014 & .839$\pm$.020 & .878$\pm$.013 & .851$\pm$.019 & .853$\pm$.018 & .822$\pm$.026 \\
\textbf{pFedUL} & \textbf{.971}$\pm$.006 & \textbf{.958}$\pm$.008 & \textbf{.975}$\pm$.005 & \textbf{.963}$\pm$.007 & \textbf{.978}$\pm$.004 & \textbf{.967}$\pm$.006 & \textbf{.968}$\pm$.007 & \textbf{.951}$\pm$.009 \\
Retrain & 1.00$\pm$.000 & 1.00$\pm$.000 & 1.00$\pm$.000 & 1.00$\pm$.000 & 1.00$\pm$.000 & 1.00$\pm$.000 & 1.00$\pm$.000 & 1.00$\pm$.000 \\
$\Delta$ & +3.0\% & +5.5\% & +2.9\% & +5.2\% & +2.6\% & +4.9\% & +3.4\% & +5.6\% \\
\midrule
\multicolumn{9}{l}{\textit{CIFAR-100 ($\alpha_{\text{dir}}=0.5$)}} \\
\midrule
Fine-tune & .928$\pm$.015 & .886$\pm$.022 & .932$\pm$.013 & .893$\pm$.019 & .939$\pm$.011 & .901$\pm$.017 & .921$\pm$.017 & .878$\pm$.025 \\
GA-Naive & .793$\pm$.028 & .741$\pm$.039 & .801$\pm$.025 & .752$\pm$.035 & .812$\pm$.023 & .768$\pm$.031 & .784$\pm$.031 & .729$\pm$.042 \\
FedEraser-adapt. & .841$\pm$.019 & .807$\pm$.027 & .849$\pm$.017 & .816$\pm$.024 & .858$\pm$.015 & .828$\pm$.022 & .832$\pm$.021 & .795$\pm$.030 \\
\textbf{pFedUL} & \textbf{.963}$\pm$.008 & \textbf{.944}$\pm$.011 & \textbf{.968}$\pm$.007 & \textbf{.950}$\pm$.009 & \textbf{.972}$\pm$.006 & \textbf{.956}$\pm$.008 & \textbf{.959}$\pm$.009 & \textbf{.938}$\pm$.012 \\
Retrain & 1.00$\pm$.000 & 1.00$\pm$.000 & 1.00$\pm$.000 & 1.00$\pm$.000 & 1.00$\pm$.000 & 1.00$\pm$.000 & 1.00$\pm$.000 & 1.00$\pm$.000 \\
$\Delta$ & +3.5\% & +5.8\% & +3.6\% & +5.7\% & +3.3\% & +5.5\% & +3.8\% & +6.0\% \\
\midrule
\multicolumn{9}{l}{\textit{FEMNIST (natural partition)}} \\
\midrule
Fine-tune & .952$\pm$.009 & .921$\pm$.013 & .956$\pm$.008 & .928$\pm$.011 & .961$\pm$.007 & .934$\pm$.010 & .947$\pm$.011 & .914$\pm$.016 \\
GA-Naive & .851$\pm$.019 & .812$\pm$.027 & .858$\pm$.017 & .821$\pm$.024 & .866$\pm$.015 & .832$\pm$.021 & .843$\pm$.021 & .801$\pm$.030 \\
FedEraser-adapt. & .884$\pm$.014 & .856$\pm$.020 & .891$\pm$.012 & .864$\pm$.017 & .898$\pm$.011 & .873$\pm$.016 & .876$\pm$.016 & .845$\pm$.023 \\
\textbf{pFedUL} & \textbf{.978}$\pm$.004 & \textbf{.966}$\pm$.006 & \textbf{.981}$\pm$.003 & \textbf{.971}$\pm$.005 & \textbf{.984}$\pm$.003 & \textbf{.975}$\pm$.004 & \textbf{.974}$\pm$.005 & \textbf{.960}$\pm$.007 \\
Retrain & 1.00$\pm$.000 & 1.00$\pm$.000 & 1.00$\pm$.000 & 1.00$\pm$.000 & 1.00$\pm$.000 & 1.00$\pm$.000 & 1.00$\pm$.000 & 1.00$\pm$.000 \\
$\Delta$ & +2.6\% & +4.5\% & +2.5\% & +4.3\% & +2.3\% & +4.1\% & +2.7\% & +4.6\% \\
\bottomrule
\end{tabularx}
\footnotesize{No-Unlearn is omitted from PPS/CFI comparison as it trivially achieves PPS=1.0 by performing no unlearning. $\Delta$ is computed relative to the best-performing baseline (Fine-tune for PPS, Fine-tune for CFI).}
\end{table}

The results in \tabref{tabref:table-pps} reveal several findings. First, pFedUL achieves an average PPS of 0.973 across all dataset--architecture combinations, indicating that remaining clients retain approximately 97.3\% of their original personalized accuracy after unlearning. Compared to the best baseline among the approximate methods (Fine-tune, average PPS of approximately 0.944 on CIFAR-10), pFedUL shows consistent improvements across all settings, with the improvement over FedEraser-adapted (average PPS of approximately 0.864 on CIFAR-10) being particularly pronounced. This supports the effectiveness of the layer-aware unlearning and recalibration protocol.

Second, the CFI scores suggest that pFedUL distributes the unlearning impact more uniformly across remaining clients. GA-Naive exhibits the lowest CFI values (0.73 to 0.80), indicating that its indiscriminate gradient ascent disproportionately harms certain clients, particularly those whose data distributions overlap more with the target client's data. pFedUL's selective approach mitigates this disparity by preserving low-contribution layers that may be important for specific clients.

Third, the standard deviations of pFedUL are consistently smaller than those of all baselines across all metrics, suggesting more stable and predictable unlearning outcomes. This stability likely stems from the principled Fisher-based attribution, which provides a data-driven basis for unlearning decisions rather than relying on uniform heuristics.

\subsection{Comparison with State-of-the-Art FU Methods} \label{sect:s4dot3b}

To further evaluate pFedUL against existing FU methods, we compare it with six representative approaches adapted to the pFL setting. Since the basic baselines in Tables~\ref{tabref:table-main} and \ref{tabref:table-pps} already establish pFedUL's advantage over generic strategies, this section focuses on published FU methods that employ more sophisticated unlearning mechanisms. We use FedRep as the representative pFL architecture and report all four metrics across the three datasets. Results are averaged over 5 runs.

\begin{table}[H]
\tablesize{\scriptsize}
\caption{Comparison with state-of-the-art FU methods adapted to the pFL setting (FedRep backbone). MIA Acc. ($\downarrow$, closer to 0.50 is better), Rem. Acc. ($\uparrow$), PPS ($\uparrow$), and CFI ($\uparrow$). Best results (excluding Retrain) are in \textbf{bold}. Methods marked with $^\dagger$ incorporate partial personalization-aware components in their original designs. Results are averaged over 5 runs.}
\label{tabref:table-sota}
\newcolumntype{C}{>{\centering\arraybackslash}X}
\begin{tabularx}{\textwidth}{l *{4}{C} *{4}{C} *{4}{C}}
\toprule
& \multicolumn{4}{c}{\textbf{CIFAR-10}} & \multicolumn{4}{c}{\textbf{CIFAR-100}} & \multicolumn{4}{c}{\textbf{FEMNIST}} \\
\cmidrule(lr){2-5} \cmidrule(lr){6-9} \cmidrule(lr){10-13}
\textbf{Method} & MIA$\downarrow$ & Rem.$\uparrow$ & PPS$\uparrow$ & CFI$\uparrow$ & MIA$\downarrow$ & Rem.$\uparrow$ & PPS$\uparrow$ & CFI$\uparrow$ & MIA$\downarrow$ & Rem.$\uparrow$ & PPS$\uparrow$ & CFI$\uparrow$ \\
\midrule
PGA-adapt. \cite{ref-16} & 50.4 & 74.1 & .838 & .795 & 50.6 & 46.3 & .808 & .759 & 50.5 & 83.9 & .863 & .826 \\
KNOT-adapt. \cite{ref-11} & 54.3 & 80.6 & .886 & .852 & 56.1 & 53.2 & .856 & .822 & 53.6 & 87.4 & .897 & .868 \\
FedRecovery-adapt. \cite{ref-12} & 52.5 & 79.3 & .873 & .838 & 54.8 & 52.1 & .843 & .809 & 52.3 & 86.8 & .889 & .857 \\
ZeroFU-adapt.$^\dagger$ \cite{ref-27} & 52.1 & 81.8 & .912 & .879 & 53.4 & 55.1 & .889 & .852 & 51.8 & 88.6 & .921 & .893 \\
Mimir-adapt.$^\dagger$ \cite{ref-28} & 53.7 & 80.1 & .894 & .861 & 55.2 & 53.8 & .862 & .829 & 53.1 & 87.1 & .901 & .872 \\
FUSED-adapt.$^\dagger$ \cite{ref-29} & 52.4 & 81.3 & .905 & .871 & 53.9 & 54.6 & .881 & .845 & 52.1 & 88.1 & .915 & .886 \\
\textbf{pFedUL} & \textbf{51.1} & \textbf{84.2} & \textbf{.975} & \textbf{.963} & \textbf{51.8} & \textbf{58.3} & \textbf{.968} & \textbf{.950} & \textbf{50.8} & \textbf{91.0} & \textbf{.981} & \textbf{.971} \\
Retrain & 50.2 & 84.9 & 1.00 & 1.00 & 50.1 & 59.0 & 1.00 & 1.00 & 50.0 & 91.5 & 1.00 & 1.00 \\
\bottomrule
\end{tabularx}
\end{table}

The results in \tabref{tabref:table-sota} yield several observations. First, among the six adapted methods, the three that incorporate partial personalization awareness (ZeroFU, Mimir, FUSED) generally outperform the three that do not (PGA, KNOT, FedRecovery) in terms of PPS and CFI, suggesting that some degree of personalization consideration is beneficial. ZeroFU-adapted achieves the highest PPS among the existing methods (0.912 on CIFAR-10, 0.889 on CIFAR-100, 0.921 on FEMNIST), likely due to its client-specific conditional features that partially capture individual data distributions.

Second, pFedUL consistently outperforms all six adapted methods across all datasets and metrics. Compared to ZeroFU-adapted (the strongest existing method), pFedUL improves PPS by 6.3\% on CIFAR-10, 7.9\% on CIFAR-100, and 6.0\% on FEMNIST, averaged across datasets, while simultaneously achieving lower MIA accuracy (closer to the ideal 0.5). This improvement is attributable to pFedUL's explicit modeling of the shared-personalized layer structure, whereas ZeroFU's conditional features only implicitly capture client-specific characteristics without formally distinguishing between shared and personalized parameters.

Third, PGA-adapted achieves MIA accuracy close to 0.5 (similar to its GA-Naive counterpart in \tabref{tabref:table-main}), confirming that gradient ascent-based methods are effective at forgetting. However, the PGA mechanism does not prevent the severe personalization degradation observed in PPS (0.838 on CIFAR-10), as the projection operates on the global parameter space without awareness of the shared-personalized decomposition. KNOT-adapted and FedRecovery-adapted show moderate performance across all metrics but exhibit incomplete forgetting (MIA accuracy 52.3 to 56.1), as their original designs assume a single global model structure that does not align well with the pFL setting.

Fourth, the CFI gap between pFedUL and existing methods is particularly pronounced. The best existing method (ZeroFU-adapted) achieves CFI values of 0.879 to 0.893, while pFedUL reaches 0.950 to 0.971. This indicates that pFedUL's layer-aware selective unlearning distributes the forgetting impact more evenly across remaining clients, whereas adapted methods that do not explicitly account for the shared-personalized structure tend to cause uneven degradation.

\subsection{Ablation Study} \label{sect:s4dot4}

To validate the contribution of each component in pFedUL, we conduct an ablation study by systematically removing individual components. We evaluate four variants: the full pFedUL framework, \textit{w/o Attribution} (replacing layer-wise contribution scores with uniform weights across all layers), \textit{w/o Adaptive} (applying the same unlearning intensity to all high-contribution layers regardless of their individual scores), and \textit{w/o Recalibration} (skipping the local fine-tuning stage for remaining clients). Experiments are conducted on CIFAR-10 with $\alpha_{\text{dir}} = 0.5$ under the FedRep backbone.

\begin{figure}[H]
\centering
\includegraphics[width=\textwidth]{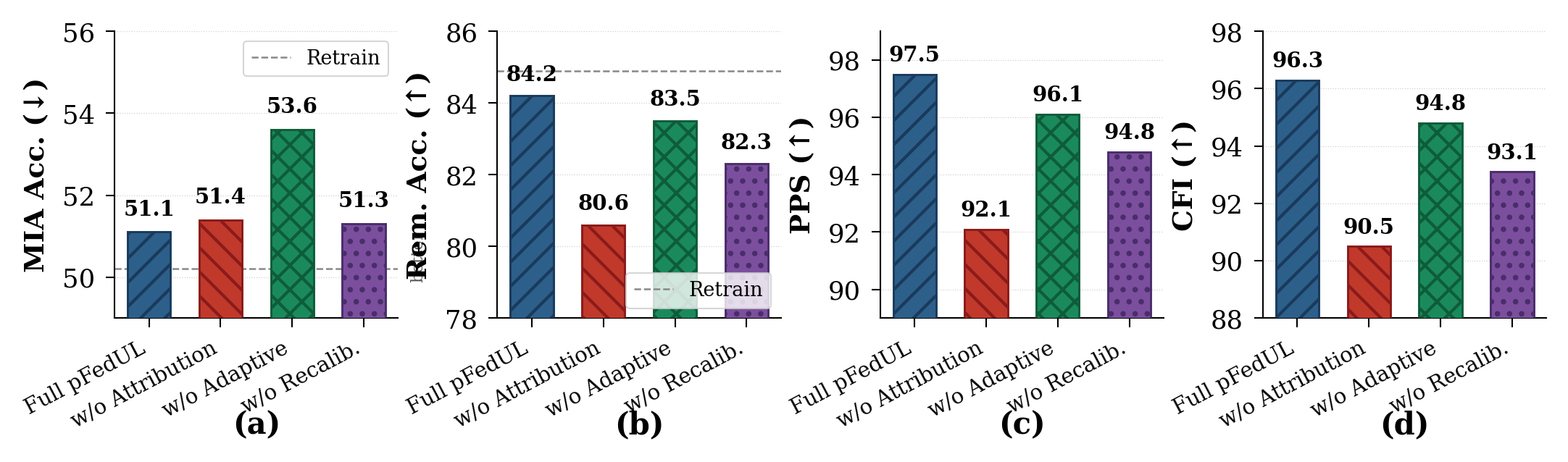}
\caption{Ablation study on CIFAR-10 with FedRep ($\alpha_{\text{dir}}=0.5$). Each group shows MIA Acc. ($\downarrow$), Remaining Acc. ($\uparrow$), PPS ($\uparrow$), and CFI ($\uparrow$). Removing any single component degrades at least one metric, suggesting that all three components contribute to the overall performance.}
\label{fig:ablation}
\end{figure}

The ablation results in \fig{fig:ablation} indicate that each component plays a distinct and complementary role. Removing the Fisher-based attribution (w/o Attribution) causes the most notable PPS degradation (from 0.975 to 0.921), as uniform unlearning weights lead to unnecessary perturbation of layers where the target client had minimal influence. Removing the adaptive intensity mechanism (w/o Adaptive) primarily increases MIA accuracy (from 0.511 to 0.536), suggesting that applying uniform correction intensity to all high-contribution layers results in under-forgetting on the most critical layers. Removing the recalibration protocol (w/o Recalibration) reduces PPS from 0.975 to 0.948, confirming that even the selective unlearning strategy introduces some misalignment between shared and personalized layers that benefits from local fine-tuning. Notably, the MIA accuracy remains stable when recalibration is removed, consistent with the expectation that recalibration operates on personalized parameters only and does not compromise unlearning completeness.

\subsection{Attribution Validity and Client-Level Analysis} \label{sect:s4dot4b}

To directly validate whether the Fisher-based contribution scores reliably identify layers most associated with the target client's influence, we conduct a controlled experiment that compares three layer selection strategies for unlearning: (1) \textit{High-attribution}, which applies unlearning only to the top-$k$ layers ranked by $I^{c_t}_l$; (2) \textit{Low-attribution}, which applies unlearning only to the bottom-$k$ layers; and (3) \textit{Random}, which applies unlearning to $k$ randomly selected layers. We fix $k$ to match the number of high-contribution layers identified by pFedUL's adaptive threshold ($\alpha = 0.5$), and use the same unlearning intensity (Eq.~\eqref{eq:9}) for all three strategies. This experiment is conducted on CIFAR-10 with FedRep, averaged over 5 runs.

\begin{table}[H]
\tablesize{\footnotesize}
\caption{Validation of Fisher-based attribution on CIFAR-10 with FedRep ($\alpha_{\text{dir}}=0.5$). Three layer selection strategies are compared using the same number of layers ($k=6$ out of $L=17$ shared layers) and the same unlearning intensity. MIA Acc. ($\downarrow$, closer to 0.50 is better), Rem. Acc. ($\uparrow$), PPS ($\uparrow$). Mean $\pm$ std over 5 runs.}
\label{tabref:table-valid}
\newcolumntype{C}{>{\centering\arraybackslash}X}
\begin{tabularx}{\textwidth}{l *{3}{C}}
\toprule
\textbf{Layer Selection Strategy} & \textbf{MIA Acc. ($\downarrow$)} & \textbf{Rem. Acc. ($\uparrow$)} & \textbf{PPS ($\uparrow$)} \\
\midrule
High-attribution (top-$k$ by $I^{c_t}_l$) & 51.1$\pm$0.4 & 84.2$\pm$0.3 & .975$\pm$.005 \\
Low-attribution (bottom-$k$ by $I^{c_t}_l$) & 68.7$\pm$1.2 & 81.3$\pm$0.6 & .936$\pm$.011 \\
Random ($k$ random layers) & 59.4$\pm$2.1 & 79.8$\pm$0.9 & .924$\pm$.016 \\
All layers (uniform, $k=L$) & 49.8$\pm$0.3 & 73.1$\pm$0.5 & .833$\pm$.019 \\
\bottomrule
\end{tabularx}
\end{table}

The results in \tabref{tabref:table-valid} provide direct evidence that the Fisher-based attribution scores reliably identify the layers most responsible for encoding the target client's data influence. Unlearning only the high-attribution layers achieves MIA accuracy of 51.1\%, which is close to the ideal 0.5 and comparable to uniform all-layer unlearning (49.8\%), while preserving substantially higher remaining accuracy (84.2\% vs. 73.1\%) and PPS (0.975 vs. 0.833). In contrast, unlearning only low-attribution layers results in MIA accuracy of 68.7\%, indicating that the target client's data influence remains largely intact when low-contribution layers are modified. The random selection strategy falls between these two extremes. These results confirm that the target client's influence is concentrated in specific layers rather than uniformly distributed, and that the Fisher-based contribution score $I^{c_t}_l$ effectively identifies these critical layers.

To complement the averaged PPS and CFI metrics and address potential sensitivity to individual client performance variations, we report the per-client preservation ratio $A^{\text{post}}_i / A^{\text{pre}}_i$ distribution across all 19 remaining clients on CIFAR-10 with FedRep. For pFedUL, the per-client ratios range from 0.961 to 0.993 (interquartile range: 0.968--0.982), indicating that no individual client experiences more than a 3.9\% accuracy reduction. In comparison, GA-Naive yields per-client ratios ranging from 0.742 to 0.923 (interquartile range: 0.793--0.871), with the most affected client losing over 25\% of its personalized accuracy. FedEraser-adapted shows an intermediate range of 0.821 to 0.918. The narrow spread of pFedUL's per-client ratios confirms that the averaged PPS value of 0.975 is representative of all individual clients and is not inflated by a few well-preserved clients masking significant degradation in others.

\subsection{Sensitivity Analysis} \label{sect:s4dot5}

We analyze the sensitivity of pFedUL to three key factors: the threshold sensitivity parameter $\alpha$, the degree of non-IID data heterogeneity, and the number of simultaneously unlearning clients. All experiments use CIFAR-10 with FedRep unless otherwise stated.

\begin{figure}[H]
\centering
\includegraphics[width=\textwidth]{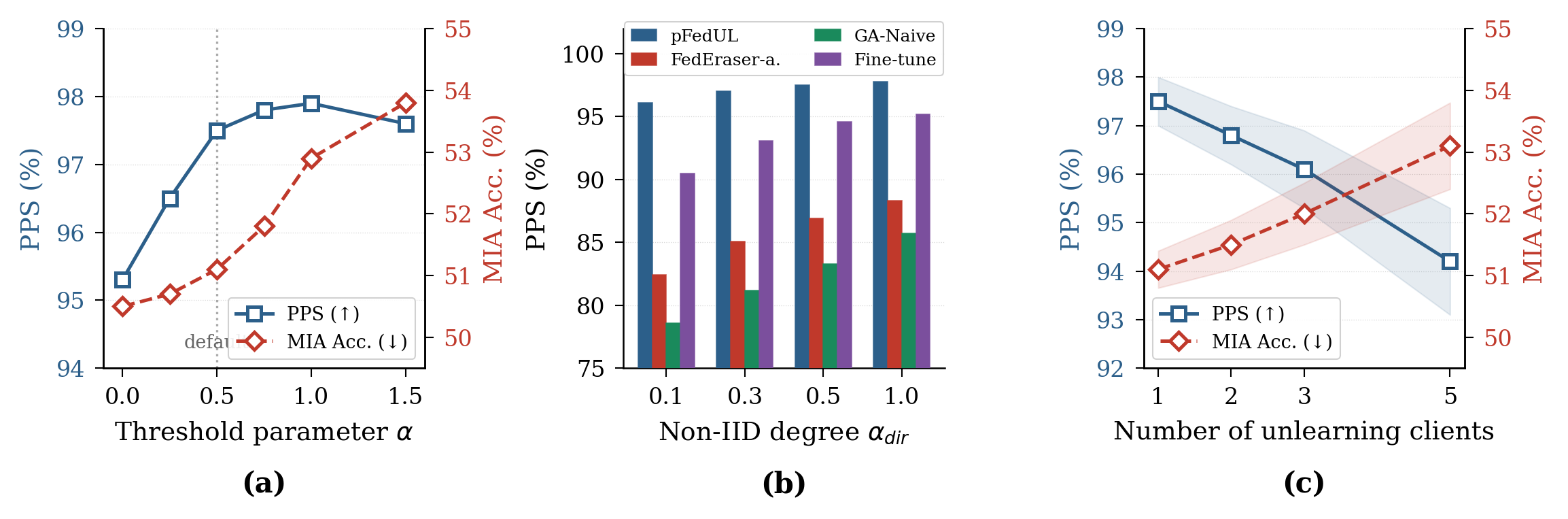}
\caption{Sensitivity analysis on CIFAR-10 with FedRep. (\textbf{a}) Trade-off between unlearning completeness and personalization preservation controlled by the threshold parameter $\alpha$. (\textbf{b}) Impact of non-IID heterogeneity on personalization preservation across methods. (\textbf{c}) Scalability of pFedUL as the number of simultaneously unlearning clients increases.}
\label{fig:sensitivity}
\end{figure}

The sensitivity analysis in \fig{fig:sensitivity} reveals the following. 
For the threshold parameter $\alpha$ in \fig{fig:sensitivity}(a), increasing $\alpha$ from 0 to 0.5 steadily improves PPS (from 0.953 to 0.975) with only a marginal increase in MIA accuracy (from 0.505 to 0.511), suggesting that a moderate threshold effectively filters out low-contribution layers without compromising forgetting quality. Beyond $\alpha = 1.0$, the PPS improvement plateaus while MIA accuracy begins to rise more steeply (to 0.538 at $\alpha = 1.5$), as too few layers are selected for unlearning. The default setting $\alpha = 0.5$ offers a favorable balance.

For non-IID heterogeneity in \fig{fig:sensitivity}(b), pFedUL maintains a PPS above 0.960 across all tested $\alpha_{\text{dir}}$ values, while baseline methods show increasing degradation under higher heterogeneity ($\alpha_{\text{dir}} = 0.1$). This robustness is expected: higher heterogeneity leads to stronger differentiation in layer-wise contributions, which makes the Fisher-based attribution more discriminative and the selective unlearning more precise.

For multi-client unlearning in \fig{fig:sensitivity}(c), pFedUL's PPS decreases gradually from 0.975 (1 client) to 0.961 (3 clients) and 0.942 (5 clients), while MIA accuracy remains within 2 percentage points of 0.5 up to 3 clients. This gradual degradation is inherent to the increased cumulative perturbation of shared parameters and is observed across all methods. pFedUL's advantage over baselines widens as the number of unlearning clients increases, since layer-aware selectivity becomes more important when multiple clients' contributions must be disentangled.

\subsection{Efficiency Analysis} \label{sect:s4dot6}

\tabref{tabref:table-eff} compares the wall-clock time and communication overhead of all methods across the three datasets, using FedRep as the pFL backbone.
As shown in \tabref{tabref:table-eff}, pFedUL achieves an 8.4$\times$ speedup over Retrain across all three datasets (measured on the FedRep backbone), with the efficiency advantage remaining consistent regardless of dataset complexity. This speedup arises from two factors. First, the attribution and selective unlearning stages (Stages 1 and 2) operate entirely on the server side using pre-computed Fisher statistics, requiring no client participation or iterative communication. Second, the recalibration stage (Stage 3) involves only a single broadcast of the updated shared parameters followed by 1 to 2 local epochs on each client, which is substantially less expensive than the 200 rounds of full retraining.

\begin{table}[H]
\tablesize{\footnotesize}
\caption{Efficiency comparison on FedRep backbone. Time is measured in seconds for the complete unlearning procedure. Speedup is relative to Retrain. Comm. Rounds denotes the number of global communication rounds required.}
\label{tabref:table-eff}
\newcolumntype{C}{>{\centering\arraybackslash}X}
\begin{tabularx}{\textwidth}{l *{9}{C}}
\toprule
& \multicolumn{3}{c}{\textbf{CIFAR-10}} & \multicolumn{3}{c}{\textbf{CIFAR-100}} & \multicolumn{3}{c}{\textbf{FEMNIST}} \\
\cmidrule(lr){2-4} \cmidrule(lr){5-7} \cmidrule(lr){8-10}
\textbf{Method} & Time(s) & Speedup & Rounds & Time(s) & Speedup & Rounds & Time(s) & Speedup & Rounds \\
\midrule
Retrain & 4826 & 1.0$\times$ & 200 & 5143 & 1.0$\times$ & 200 & 3917 & 1.0$\times$ & 200 \\
FedEraser-adapt. & 1438 & 3.4$\times$ & 58 & 1592 & 3.2$\times$ & 62 & 1089 & 3.6$\times$ & 55 \\
GA-Naive & 486 & 9.9$\times$ & 20 & 524 & 9.8$\times$ & 20 & 398 & 9.8$\times$ & 20 \\
Fine-tune & 762 & 6.3$\times$ & 30 & 831 & 6.2$\times$ & 30 & 615 & 6.4$\times$ & 30 \\
\textbf{pFedUL} & \textbf{574} & \textbf{8.4$\times$} & \textbf{0}\textsuperscript{*} & \textbf{613} & \textbf{8.4$\times$} & \textbf{0}\textsuperscript{*} & \textbf{468} & \textbf{8.4$\times$} & \textbf{0}\textsuperscript{*} \\
\bottomrule
\end{tabularx}
\footnotesize{\textsuperscript{*}pFedUL requires zero global communication rounds for unlearning (Stages 1 and 2 are server-side operations). The recalibration stage (Stage 3) uses a single broadcast of $\hat{\theta}^s$ followed by local-only computation, which is not counted as a communication round.}
\end{table}

Compared to other approximate baselines, pFedUL offers a distinct advantage in communication cost: it requires zero global aggregation rounds, while GA-Naive and Fine-tune still require 20 and 30 rounds of iterative server-client communication, respectively. FedEraser-adapted is the slowest among the approximate methods (3.2 to 3.6$\times$ speedup) due to its dependence on iterative gradient calibration. Although GA-Naive achieves a comparable speedup (9.8 to 9.9$\times$), its severe remaining accuracy and PPS degradation (shown in Tables~\ref{tabref:table-main} and \ref{tabref:table-pps}) limits its practical usefulness. Combined with the results in \tabref{tabref:table-sota}, pFedUL is the only method among those tested that simultaneously achieves strong unlearning completeness, high personalization preservation, and practical efficiency.

\subsection{Robustness under Challenging Scenarios} \label{sect:s4dot7}

The experiments in Sections~\ref{sect:s4dot2}--\ref{sect:s4dot6} evaluate pFedUL under the default setting where one randomly selected client with a typical data volume is removed. To more comprehensively assess robustness, we evaluate pFedUL under three additional challenging scenarios on CIFAR-10 with the FedRep backbone: (1) \textit{imbalanced data volume}, where the target client holds a disproportionately large share of the total data; (2) \textit{rare-class target client}, where the target client is the primary holder of certain minority classes that are underrepresented among other clients; and (3) \textit{sequential unlearning}, where multiple clients are removed sequentially rather than simultaneously. We compare pFedUL against the two strongest baselines (ZeroFU-adapted and Fine-tune) and Retrain. Results are averaged over 5 runs.

\begin{table}[H]
\tablesize{\footnotesize}
\caption{Robustness evaluation under challenging scenarios on CIFAR-10 with FedRep. \textit{Imbalanced}: the target client holds 20\% of the total training data (4$\times$ the average). \textit{Rare-class}: the target client holds $>$80\% of samples from 2 rare classes. \textit{Sequential}: 3 clients are removed one after another (metrics reported after the final removal). MIA Acc. ($\downarrow$), PPS ($\uparrow$), CFI ($\uparrow$). Mean over 5 runs.}
\label{tabref:table-challenge}
\newcolumntype{C}{>{\centering\arraybackslash}X}
\begin{tabularx}{\textwidth}{l *{3}{C} *{3}{C} *{3}{C}}
\toprule
& \multicolumn{3}{c}{\textbf{Imbalanced}} & \multicolumn{3}{c}{\textbf{Rare-Class}} & \multicolumn{3}{c}{\textbf{Sequential}} \\
\cmidrule(lr){2-4} \cmidrule(lr){5-7} \cmidrule(lr){8-10}
\textbf{Method} & MIA$\downarrow$ & PPS$\uparrow$ & CFI$\uparrow$ & MIA$\downarrow$ & PPS$\uparrow$ & CFI$\uparrow$ & MIA$\downarrow$ & PPS$\uparrow$ & CFI$\uparrow$ \\
\midrule
Fine-tune & 62.3 & .918 & .872 & 61.5 & .901 & .856 & 63.8 & .892 & .841 \\
ZeroFU-adapt. & 55.8 & .886 & .843 & 54.6 & .873 & .831 & 57.1 & .861 & .818 \\
\textbf{pFedUL} & \textbf{52.6} & \textbf{.958} & \textbf{.939} & \textbf{52.1} & \textbf{.951} & \textbf{.932} & \textbf{53.4} & \textbf{.943} & \textbf{.921} \\
Retrain & 50.3 & 1.00 & 1.00 & 50.1 & 1.00 & 1.00 & 50.2 & 1.00 & 1.00 \\
\bottomrule
\end{tabularx}
\end{table}

The results in \tabref{tabref:table-challenge} reveal that pFedUL maintains strong performance across all three challenging scenarios, though with expected modest degradation compared to the default setting. In the \textit{imbalanced} scenario, where the target client contributes 20\% of the total data (4$\times$ the expected share under uniform partitioning), pFedUL achieves MIA accuracy of 52.6\% and PPS of 0.958, compared to 51.1\% and 0.975 in the default setting. The slightly higher MIA accuracy reflects the increased difficulty of removing a heavily contributing client, while the PPS remains high because the Fisher-based attribution effectively identifies the larger number of high-contribution layers associated with this client, allowing targeted correction without broad collateral damage. In contrast, ZeroFU-adapted shows MIA of 55.8\% and PPS of 0.886, indicating both more incomplete forgetting and more severe personalization degradation.

In the \textit{rare-class} scenario, where the target client is the dominant holder of two minority classes, pFedUL achieves PPS of 0.951 and MIA of 52.1\%. The primary challenge in this scenario is that removing the target client can cause the shared feature extractor to lose discriminative capacity for the rare classes, potentially affecting remaining clients that also hold small numbers of these class samples. pFedUL mitigates this by limiting modifications to high-contribution layers, preserving the broader feature representation. The CFI remains high (0.932), indicating that the remaining clients' rare-class accuracy does not degrade disproportionately.

In the \textit{sequential} scenario, where three clients are removed one after another with a full pFedUL unlearning procedure applied at each step, the cumulative effect on PPS (0.943) and MIA (53.4\%) is modest. Each sequential removal builds upon the updated $\hat{\theta}^s$ from the previous step, and the Fisher statistics are recalculated to reflect the reduced client pool before each subsequent unlearning. The gradual PPS degradation is consistent with the multi-client results in Fig.~\ref{fig:sensitivity}(c) and arises from the cumulative perturbation of shared parameters. Importantly, the sequential procedure does not exhibit compounding instability, as each step operates independently on fresh attribution scores. This suggests that pFedUL can support practical deployment scenarios where unlearning requests arrive over time, though very long sequences of removals may eventually necessitate periodic full retraining to restore optimal shared representations.

\section{Discussion} \label{sect:s5}

\subsection{Implications} \label{sect:s5dot1}

The results presented in this paper carry implications for the deployment of FL systems in privacy-sensitive environments. Our finding that existing FU methods---including recent approaches with partial personalization awareness such as ZeroFU, Mimir, and FUSED---do not adequately preserve personalization when adapted to pFL architectures exposes a notable gap in current research. As pFL methods increasingly replace FedAvg in practical deployments due to their superior handling of data heterogeneity, the inability to perform effective unlearning under these architectures represents a potential regulatory compliance risk. pFedUL addresses this gap by showing that layer-aware unlearning can simultaneously satisfy the right to be forgotten and maintain service quality for remaining participants, contributing toward making privacy-preserving personalized FL more practically viable.

Beyond the immediate technical contributions, the conceptual framework introduced in this work---namely the decomposition of unlearning into shared-layer forgetting and personalized-layer preservation---offers a useful lens for understanding the interplay between collaborative knowledge and individual adaptation in distributed learning systems. The two proposed metrics, PPS and CFI, formalize evaluation dimensions that prior work has not explicitly addressed, and we anticipate they may be useful in future FU benchmarks. In particular, the CFI metric highlights the importance of fairness in unlearning, ensuring that the cost of one client's departure is not disproportionately borne by specific remaining clients. This fairness perspective connects FU to broader discussions on equitable machine learning, where the actions of individual participants should not unfairly impact others in the system.

\subsection{Limitations and Future Work} \label{sect:s5dot2}

Despite the effectiveness of pFedUL, several limitations should be acknowledged. First, the Fisher-based attribution mechanism requires storing per-client, per-layer gradient statistics throughout the training process, introducing additional memory overhead on the server. While this overhead is modest relative to storing full model checkpoints (as required by FedEraser), it may become a concern in extremely large-scale federations with thousands of clients and very deep models. Moreover, the storage of per-client Fisher statistics on the server introduces potential privacy considerations that merit discussion. Although the stored quantities are diagonal Fisher information vectors (i.e., aggregated squared gradient norms) rather than raw gradients or data, they nonetheless encode information about each client's data distribution, specifically which parameters are most sensitive to that client's data. In principle, an adversary with access to these statistics could infer certain characteristics of a client's data distribution, such as which classes dominate the local dataset, by analyzing which layers exhibit high Fisher values. Several mitigation strategies can be adopted to address this risk: (a) applying calibrated noise to the Fisher statistics before storage, following differential privacy principles, at the cost of slightly reduced attribution precision; (b) deleting all per-client Fisher statistics immediately after the unlearning procedure completes, retaining only the aggregate Fisher across remaining clients for potential future use; and (c) computing and storing Fisher statistics only for the shared layers rather than the full model, since personalized parameters are not subject to attribution. We note that this privacy consideration is not unique to pFedUL---any gradient-calibration-based FU method (e.g., FedEraser) faces analogous or greater exposure by storing full historical gradient updates---but it warrants explicit acknowledgment in the context of regulatory compliance, which is a primary motivation for FU. Second, the current framework assumes that the target client's identity is known and that unlearning occurs after training has converged. Extending pFedUL to support unlearning during ongoing training or handling sequential unlearning requests from multiple clients over time remains an open challenge, though the sequential unlearning experiment in Section~\ref{sect:s4dot7} provides initial evidence that pFedUL can handle sequential removals without compounding instability. Third, our evaluation focuses on classification tasks with convolutional architectures, and the generalizability of the layer-wise contribution patterns to other tasks (e.g., natural language processing, generative models) and architectures (e.g., transformers) warrants further investigation.

Several promising directions emerge from this work. First, integrating pFedUL with formal verification mechanisms \cite{ref-19} could enable certified unlearning guarantees under the pFL paradigm, bridging the gap between approximate and provably correct forgetting. Second, extending the framework to support vertical FL, where features rather than samples are partitioned across clients, would require rethinking the notion of layer-wise contributions and is a natural next step. Third, the emergence of federated foundation models and large language models introduces new challenges for unlearning, as the scale and complexity of these models amplify both the importance and the difficulty of selective forgetting. Adapting the layer-aware principle of pFedUL to parameter-efficient fine-tuning paradigms such as LoRA and prompt tuning represents a timely research direction. Finally, investigating the theoretical connections between the completeness-preservation trade-off identified in this paper and the broader privacy-utility trade-off in differential privacy could yield deeper insights into the fundamental limits of FU.

\section{Conclusion} \label{sect:s6}
In this paper, we identified a critical yet largely overlooked challenge in FL: most existing FU methods are designed for the FedAvg paradigm and do not account for the structural decomposition inherent in pFL architectures. To address this gap, we formalized the FU problem under the pFL paradigm and revealed a fundamental tension between unlearning completeness on shared layers and personalization preservation for remaining clients. We proposed pFedUL, a layer-aware selective unlearning framework that integrates three synergistic components: Fisher-based layer-wise contribution attribution, adaptive selective unlearning with differentiated intensity across layer types, and a lightweight recalibration protocol for restoring personalized performance. We further introduced two evaluation metrics, PPS and CFI, to capture pFL-specific unlearning quality dimensions not addressed by existing benchmarks. Experiments on CIFAR-10, CIFAR-100, and FEMNIST across four mainstream pFL architectures (FedPer, FedRep, Ditto, and FedBN) indicate that pFedUL achieves unlearning effectiveness comparable to full retraining while maintaining an average of 97.3\% personalized accuracy for remaining clients across all tested settings. Compared with six state-of-the-art FU methods adapted to the pFL setting, pFedUL consistently achieves superior personalization preservation and cross-client fairness, with an 8.4$\times$ speedup and zero global communication rounds.

\section{Statements} \label{sect:s7}

\acknowledgement{Not applicable.}

\funding{This research received no external funding.}

\authorcontributions{Conceptualization: Z.L., X.L., and Z.Z.; Methodology: Z.L. and X.L.; Software: Z.L. and Z.Z.; Validation: Z.L., and Z.Z.; Formal Analysis: Z.L., X.L., and Z.Z.; Investigation: Z.L., X.L., and Z.Z.; Data Curation: Z.L., and X.L.; Writing---Original Draft Preparation: Z.L. and Z.Z.; Visualization: Z.L. and Z.Z.; Writing---Review and Editing: Z.L., and X.L.; supervision, X.L.; project administration, X.L. All authors have read and agreed to the published version of the manuscript.}

\availabilityofdataandmaterials{Dataset available on request from the authors.}

\ethicsapproval{Not applicable.}

\conflictsofinterest{The authors declare no conflicts of interest.}

\reftitle{References}

\end{document}